\documentclass[11pt]{article}

\usepackage[preprint]{acl}

\usepackage{times}
\usepackage{latexsym}
\usepackage[T1]{fontenc}
\usepackage[utf8]{inputenc}
\usepackage{microtype}
\usepackage{inconsolata}
\usepackage{graphicx}
\usepackage{xcolor}
\usepackage{url}
\usepackage{multirow}
\usepackage{threeparttable}
\usepackage{booktabs}
\usepackage{tabularx}
\usepackage{makecell}
\usepackage{array}
\usepackage{siunitx}
\usepackage{enumitem}
\usepackage[whole]{bxcjkjatype}

\title{What Makes Decoder-Based Sentence Embeddings Strong? \\ Rethinking Encoder and Decoder Architectures for Sentence Embeddings}

\author{Tatsuki Okada$^{1}$ \quad Toru Sugimoto$^{2}$ \quad Yusuke Sakai$^{1}$ \quad Taro Watanabe$^{1}$ \\
  $^{1}$Nara Institute of Science and Technology (NAIST), Japan \\
  $^{2}$Shibaura Institute of Technology, Japan \\
  \texttt{okada.tatsuki.ou0@naist.ac.jp, sugimoto@shibaura-it.ac.jp} \\
  \texttt{\{sakai.yusuke.sr9, taro\}@is.naist.jp}}

\begin{document}

\maketitle
\begin{abstract}

It is known that adapting large pre-trained decoder models for sentence embeddings achieves strong performance. However, it remains unclear whether this advantage comes from the decoder itself or from substantially larger pre-training. In this study, we systematically align encoder and decoder pairs with identical parameter counts, tokenizers, training data, and training procedures, and evaluate them in Japanese and English at 0.5B and 1B parameters.
Under these controlled settings, the randomly initialized encoder consistently outperforms the randomly initialized decoder on average across all languages, model sizes, and decoder implementations, within pre-training budgets of up to 20B tokens. In contrast, a decoder pre-trained on 10T tokens surpasses the encoder when adapted with the same training recipe. These results provide no evidence that decoders are inherently superior for sentence embeddings. Instead, their advantage largely stems from transferred pre-training, and adapting a model pre-trained on 100B tokens achieves performance comparable to adapting another pre-trained one on 18T tokens, suggesting that pre-training volume alone does not fully explain the gain. Furthermore, we reveal that sentence-level language understanding, rather than span extraction, is more closely associated with embedding quality.
Our findings suggest a simple guideline: train an encoder when starting from scratch, and adapt a pre-trained model when available.
\end{abstract}

\section{Introduction}
\begin{figure}[t!]
  \centering
  \includegraphics[width=0.992\linewidth]{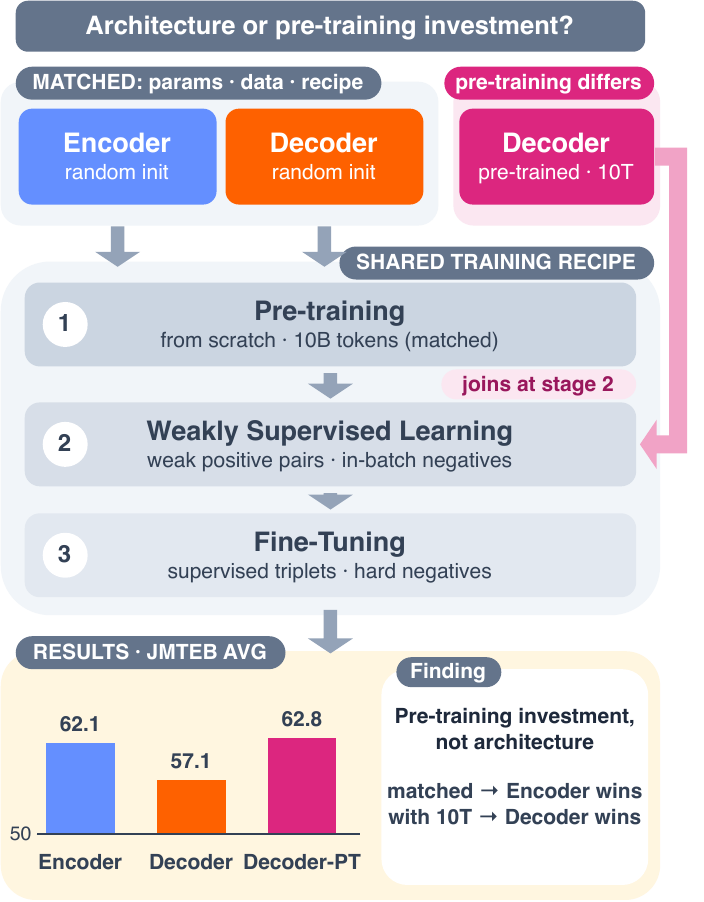}
  \caption{Overview of this study. Encoder and decoder models trained under identical conditions, including parameter count, tokenizer, training data, and training procedure, are compared with an adapted pre-trained decoder that differs only in the pre-training data, using the same sentence embedding model construction pipeline.}
  \label{concept}
\end{figure}

Sentence embeddings are a fundamental component of many natural language processing applications, such as information retrieval. By mapping sentences into dense vectors, they enable efficient similarity computation for semantic retrieval and instruction-aware language understanding. They also serve as the foundation of retrieval-augmented generation (RAG) \cite{rag}, which retrieves relevant evidence from an external corpus before generating a response with large language models (LLMs).
Consequently, the quality of sentence embeddings directly affects the performance of these applications, motivating research on improving sentence embedding models.

Encoder-based models such as BERT \cite{bert, sbert} have long been the standard architecture for sentence embeddings. More recently, decoder-based methods have achieved strong performance by adapting large pre-trained language models, e.g., SGPT~\cite{sgpt}, LLM2Vec~\cite{llm2vec}, and others~\cite{nv-embed, llama2vec, e5mistral}. Their success is generally attributed to knowledge transferred from large-scale pre-training. However, these methods substantially differ from conventional encoders in architecture, parameter count, pre-training corpus, and training procedure, making it difficult to quantify the source of the improvement. Recent studies have begun to compare matched encoder-decoder pairs trained under identical conditions \cite{ettin, mlmvsclm}. However, these studies primarily focus on the language model pre-training stage rather than sentence embedding models. Consequently, they neither evaluate the standard two-stage contrastive training pipeline for sentence embeddings nor isolate the effect of adapting a large pre-trained decoder under the same training recipe.

Therefore, the reported success of decoder-based sentence embeddings still confounds two factors: the model family, i.e., encoders pre-trained with masked language modeling (MLM) versus decoders pre-trained with causal language modeling (CLM), and knowledge transferred through substantially larger pre-training. In this study, as illustrated in Figure~\ref{concept}, we disentangle these factors by training matched encoder-decoder pairs under identical conditions and comparing their downstream performance, computational efficiency, and representation quality. We further adapt a large pre-trained decoder under the same training recipe to isolate the effect of transferred pre-training. Our experiments span two languages, Japanese and English, two model sizes, 0.5B and 1B parameters, multiple decoder architectures, and multiple pre-training sources. Our contribution lies in this controlled comparison, which examines whether decoder-based sentence embedding models retain their advantage over encoder-based models when pre-training conditions are matched.
The main findings of our experiments and analyses are as follows:

\begin{itemize}[leftmargin=*, itemsep=0em, topsep=0.5em, left=0.5em]
  \item \textbf{Under identical conditions, randomly initialized encoders consistently outperform randomly initialized decoders.} Across all languages, model sizes, and decoder architectures we evaluate (i.e., Llama, Mistral, and Qwen2.5), no randomly initialized decoder outperforms the encoder on average. For the English 0.5B models, this performance ordering holds across all pre-training budgets evaluated.
  \item \textbf{The advantage of decoder-based sentence embeddings primarily comes from transferred pre-training.} A decoder pre-trained on 10T tokens surpasses the encoder when adapted with the same training recipe. In addition, adapting a model pre-trained on 100B tokens achieves performance comparable to one pre-trained on 18T tokens, suggesting that pre-training volume alone does not explain the improvement.
  \item \textbf{Sentence-level language understanding is more closely associated with embedding quality than span extraction.} Semantic similarity, inference, and commonsense reasoning consistently align with embedding performance, whereas span extraction does not.
\end{itemize}
We release all model checkpoints\footnote{\url{https://hf.co/collections/iamtatsuki05/mirei}} and source code\footnote{\url{https://github.com/iamtatsuki05/MIREI}} under the MIT License to support future studies on sentence embedding models.

\section{Background and Related Work}
\label{sec:related}

\paragraph{Sentence embeddings.}

Modern sentence embedding methods mainly follow two directions. The first trains encoder models for sentence embeddings. Early methods, e.g., Sentence-BERT~\cite{sbert}, fine-tune pre-trained encoders, e.g., BERT \cite{bert}, using sentence pairs. More recent methods adopt contrastive learning together with weakly supervised learning, including E5 \cite{e5}, GTE \cite{gte}, and C-Pack \cite{cpack}, which have become the standard training recipe. In Japanese, Ruri \cite{ruri} follows the same recipe and provides corresponding datasets and models.
The second direction adapts large pre-trained decoder models for sentence embeddings \cite{sgpt, e5mistral, nv-embed, llama2vec}. Rather than training a sentence encoder from scratch, these methods leverage knowledge acquired during large-scale language model pre-training. For instance, LLM2Vec \cite{llm2vec} converts decoder attention into bidirectional attention and subsequently adapts the model with contrastive learning for sentence embedding.

\paragraph{Encoder vs. decoder.}
Despite the strong empirical performance of decoder-based methods, existing studies do not isolate the effect of the model family. Encoder and decoder models typically differ in parameter count, training settings, and pre-training data, making direct comparisons difficult. Moreover, language model performance follows scaling laws relating model size, training data, and compute \cite{kaplan2020scaling, chinchilla}. As a result, differences in pre-training scale can substantially affect downstream performance, highlighting the importance of controlled comparisons that disentangle the effect of the model family from transferred pre-training.
Some studies have attempted such controlled comparisons. Ettin \cite{ettin} released matched encoder-decoder pairs trained on identical data with the same pre-training recipe, showing that encoders outperform decoders on classification and retrieval, and that cross-objective continued pre-training on 50B tokens does not close the gap.
\citet{mlmvsclm} compared masked and causal language modeling across 38 models, finding that masked language modeling produces better representations while explicitly excluding contrastive post-training. ModernGBERT \cite{moderngbert} compared scratch-trained German encoders with LLM2Vec-based decoders. However, the compared models differ in size, and the scratch and adapted models share the same pre-training corpus. Our study complements these prior studies by applying both stages of contrastive sentence embedding training to all models, isolating the effect of transferred pre-training, and conducting experiments in both Japanese and English.

\section{Methods}
\label{sec:setup}

\paragraph{Overview.}

As illustrated in Figure~\ref{concept}, we compare encoders and decoders for sentence embeddings while controlling the non-embedding parameter count, tokenizer, training data, and optimization settings. \footnote{Note that, throughout this paper, \textit{encoder} and \textit{decoder} refer to model families pre-trained with MLM and CLM, respectively.
Since decoder-based models use bidirectional attention during both sentence embedding training and inference, the two model families differ in their pre-training objectives and architectural design choices within Transformer blocks, such as grouped-query attention and local attention, which are not independently controlled in our experiments. Instead, we compare multiple decoder architectures to assess the extent to which our findings depend on these architectural differences in Section~\ref{sec:decoder-architecture}.
}
Specifically, we compare three models that share the same model configuration: a randomly initialized encoder, a randomly initialized decoder, and an adapted pre-trained decoder. First, we pre-train the randomly initialized encoder and decoder from scratch using their corresponding language modeling objectives. We then adapt all models to sentence embeddings using the same two-stage contrastive training pipeline. Finally, we evaluate the resulting sentence embedding models on downstream tasks and compare their performance.

\paragraph{Comparison.}

We compare models with 0.5B and 1B parameters in Japanese and English. These settings cover two typologically different languages while remaining feasible for controlled pre-training from scratch. We use the Sarashina2.2 tokenizer\footnote{\url{https://hf.co/collections/sbintuitions/sarashina22}\label{fn:sarashina}} throughout the main comparison because it supports both Japanese and English. Using a single tokenizer eliminates differences in tokenization as a confounding factor. The encoder and decoder of each model size also share the same vocabulary and the hidden size, ensuring that their embedding matrices have identical shapes and that their output dimensions match. Further details are provided in Appendix~\ref{appendix:training-settings}.
Following the Chinchilla scaling law~\cite{chinchilla}, we set the pre-training data budget to 20 tokens per non-embedding parameter, corresponding to about 10B tokens for 0.5B models and about 20B tokens for 1B models.
Note that MLM and CLM differ in supervision density even under the same token budget. MLM computes the loss only on masked tokens, whereas CLM computes the loss at every token. Since this difference is inherent to the two training objectives, we retain it while controlling all other factors, including the training data and token budget.

\subsection{Model Configurations}
\label{sec:model-configurations}

We build and evaluate three models for each language and model size. We investigate two languages, Japanese and English, and two model sizes, 0.5B and 1B parameters, a total of 12 models.

\begin{description}[leftmargin=*, itemsep=0em, topsep=0.5em, left=0.15em]
\item[Randomly initialized encoder.] We adopt the architecture and hyperparameter configuration of ModernBERT \cite{modernbert}. The model is pre-trained from scratch with MLM and then adapted for sentence embeddings using the standard two-stage contrastive training pipeline.

\item[Randomly initialized decoder.] We adopt the architecture and hyperparameter configuration of Llama \cite{llama}. The model is pre-trained from scratch with CLM and then adapted for sentence embeddings. Following LLM2Vec \cite{llm2vec}, we convert the causal attention mechanism to bidirectional attention during sentence embedding training. 
Note that we do not incorporate masked next-token prediction (MNTP), an additional adaptation stage introduced in LLM2Vec, because it would add a language modeling stage only for the decoder, thereby compromising the comparability of the training budgets.
Instead, we apply contrastive learning directly after enabling bidirectional attention, following common practice \cite{nv-embed}, as noted by \citet{bidirlm}.

\item[Pre-trained decoder.] We use the Sarashina2.2 model series\textsuperscript{\ref{fn:sarashina}}, including the 0.5B and 1B checkpoints. These models have been pre-trained on roughly 10T Japanese and English tokens. We convert the attention mechanism to bidirectional attention following LLM2Vec and then adapt the models for sentence embeddings using the same training pipeline as the randomly initialized models. For the English setting, we use the same pre-trained checkpoint and change only the sentence embedding training data to English.
\end{description}

\subsection{Training Corpora}
\label{sec:datasets}

Training consists of two phases: language model pre-training and sentence embedding training.

Language model pre-training also consists of two stages. For Japanese, we first pre-train on FineWeb-2-Edu-Japanese \cite{penedo2025fineweb} with a sequence length of 1,024 and then continue pre-training on Wiki40B Japanese \cite{wiki40b} with a sequence length of 8,192. The total pre-training data amounts to about 10B tokens for 0.5B models and about 20B tokens for 1B models.
Sentence embedding training also consists of two stages: weakly supervised learning (WSL) on ruri-dataset-v2-pt \cite{ruri}, followed by supervised fine-tuning (FT) on ruri-v3-dataset-ft \cite{ruri}.

The English models follow the same two-stage pre-training and sentence embedding training pipeline. We use FineWeb-Edu \cite{fineweb} and Wiki40B English with matched token budgets for pre-training. For sentence embedding training, we use English pairs and triplets from embedding-training-data, including AllNLI~\cite{snli, williams-etal-2018-broad}, MS MARCO~\cite{msmarco}, Quora, and QA datasets. These English datasets serve as the source datasets for the Japanese supervision data used to construct the Ruri training corpus. Additional details are provided in Appendix~\ref{appendix:training-settings}, and links to all corpora are listed in Appendix~\ref{appendix:licenses}.

\subsection{Training Setting}
\label{sec:training-setting}

For language model pre-training, we use the naive implementations of MLM and CLM provided by the Transformers Trainers \cite{transformerslib}.

Sentence embedding training follows the standard two-stage contrastive learning pipeline. Both stages optimize the InfoNCE objective \cite{infonce}. Weakly supervised learning uses large-scale weakly supervised sentence pairs with in-batch negatives. Supervised fine-tuning uses supervised pairs and triplets together with explicit hard negatives. Each stage runs for one epoch, with a total batch size of 4,096 for weakly supervised learning and 512 for supervised fine-tuning, and each batch contains samples from a single dataset \cite{gte}.
Unless otherwise noted, we follow the training hyperparameter settings of  ModernBERT \cite{modernbert}, Chinchilla \cite{chinchilla}, and E5 \cite{e5}. The overall training budget, maximum sequence length, and optimization settings are kept identical across models of the same size. Only the learning rate is adjusted according to model size.
For all models, sentence embeddings are obtained by mean pooling. Note that we do not use instructions or textual prefixes \cite{instructor} during training or evaluation. Embeddings are L2-normalized at evaluation time, and similarity is computed using cosine similarity.
Appendix~\ref{appendix:training-settings} provides detailed training settings and model configurations.

\section{Experimental Setup}
\label{sec:evaluation}

\paragraph{Benchmarks.}
We evaluate Japanese models on JMTEB v2 \cite{jmteb}. The benchmark covers five task families: \textit{Retrieval}, \textit{STS}, \textit{Classification}, \textit{Reranking}, and \textit{Clustering} across 28 datasets. We evaluate English models on all MTEB v2 English tasks \cite{mteb, mmteb} belonging to the same five task families. This results in 37 evaluation tasks in total. The complete task list is provided in Appendix~\ref{appendix:mteb-tasks}. We report macro-averages over the five task families. %

\paragraph{Efficiency and representation quality.}
To evaluate inference efficiency, we measure throughput and peak GPU memory usage while encoding Wiki40B texts at sequence lengths of 128 and 512 tokens. We evaluate only Japanese models because the English models share the same architectures. We disable the decoder's key-value cache because the cached states are not reused during sentence embedding inference. Detailed settings are provided in Appendix~\ref{appendix:efficiency}.
To evaluate representation quality, we measure \textit{\textbf{alignment}} and \textit{\textbf{uniformity}} \cite{understanding}, which are widely used to analyze contrastive sentence embedding models such as SimCSE \cite{simcse}. Alignment measures the distance between positive sentence pairs, whereas uniformity measures how uniformly sentence embeddings are distributed on the unit hypersphere. Lower values indicate better representations for both metrics. We sample positive pairs from Wikipedia and MIRACL \cite{miracl} for Japanese, and from Wiki40B English and AllNLI for English. We use bfloat16 with a maximum sequence length of 8,192.

\paragraph{Language understanding.}
To evaluate the language understanding ability of each pre-trained model before sentence embedding training, we fine-tune the models on five Japanese tasks: JSTS, JNLI, JSQuAD, JCommonsenseQA, and JCoLA~\cite{jglue, jcola}. Following \citet{sakai2024templates}, we use the corresponding English tasks: STS-B \cite{cer-etal-2017-semeval}, MNLI \cite{williams-etal-2018-broad}, SQuAD \cite{squad}, CommonsenseQA~\cite{commonsenseqa}, and CoLA~\cite{cola}. Detailed settings are provided in Appendix~\ref{appendix:diag}.

\section{Results}
\label{sec:results}

\begin{table}[t]
  \centering
  \renewcommand{\arraystretch}{1.2}
  \small
  \setlength{\tabcolsep}{2pt}
  \begin{tabular}{@{}llcccccc@{}}
    \toprule
    & \bfseries Model & \bfseries Avg. & \bfseries Retr. & \bfseries STS & \bfseries Class. & \bfseries Rerank. & \bfseries Clust. \\
    \midrule
    \multirow{3}{*}{\rotatebox[origin=c]{90}{\footnotesize JP 0.5B}}
    & Encoder & 62.07 & \bfseries 61.11 & 80.61 & 55.38 & 65.11 & 48.15 \\
    & Decoder & 57.13 & 50.17 & 77.67 & 52.96 & 56.99 & 47.85 \\
    & Decoder-PT & \bfseries 62.81 & 55.47 & \bfseries 83.44 & \bfseries 58.50 & \bfseries 66.82 & \bfseries 49.80 \\
    \midrule
    \multirow{3}{*}{\rotatebox[origin=c]{90}{\footnotesize JP 1B}}
    & Encoder & 63.51 & 60.52 & 81.61 & 57.92 & 67.04 & 50.47 \\
    & Decoder & 61.50 & 56.38 & 79.14 & 55.63 & 64.66 & \bfseries 51.70 \\
    & Decoder-PT & \bfseries 66.36 & \bfseries 65.03 & \bfseries 83.16 & \bfseries 61.90 & \bfseries 71.37 & 50.34 \\
    \midrule
    \multirow{3}{*}{\rotatebox[origin=c]{90}{\footnotesize EN 0.5B}}
    & Encoder & 53.66 & 40.32 & 76.04 & 67.31 & 42.93 & 41.69 \\
    & Decoder & 52.12 & 37.31 & 74.10 & 66.28 & 42.29 & 40.63 \\
    & Decoder-PT & \bfseries 56.54 & \bfseries 42.62 & \bfseries 78.91 & \bfseries 72.45 & \bfseries 45.39 & \bfseries 43.35 \\
    \midrule
    \multirow{3}{*}{\rotatebox[origin=c]{90}{\footnotesize EN 1B}}
    & Encoder & 54.77 & 42.27 & 76.72 & 68.53 & 43.87 & 42.46 \\
    & Decoder & 52.93 & 39.04 & 75.45 & 66.42 & 42.47 & 41.28 \\
    & Decoder-PT & \bfseries 57.56 & \bfseries 44.90 & \bfseries 79.78 & \bfseries 73.49 & \bfseries 45.98 & \bfseries 43.64 \\
    \bottomrule
  \end{tabular}
  \caption{Sentence embedding results on JMTEB v2 for Japanese and the corresponding English MTEB tasks. \textit{``Encoder''} denotes a randomly initialized encoder, \textit{``Decoder''} denotes a randomly initialized decoder, and \textit{``Decoder-PT''} denotes the adapted pre-trained decoder. Results are grouped by language and model size. ``Avg.'' denotes the macro-average across the five task families. \textit{``Retr.''}, \textit{``Class.''}, \textit{``Rerank.''}, and \textit{``Clust.''} denote Retrieval, Classification, Reranking, and Clustering, respectively. Bold values indicate the best result within each setting. Henceforth, we use the same notation.}
  \label{tab:main}
\end{table}

\subsection{Sentence Embedding Performance}
\label{sec:Sentence-Embedding-Performance}

Table~\ref{tab:main} shows the results of the sentence embedding performance on JMTEB and the corresponding subset of English MTEB tasks.

\paragraph{Finding 1: Under identical conditions, the encoder consistently outperforms the decoder.}
For instance, for the Japanese 0.5B models, the randomly initialized encoder outperforms the matched randomly initialized decoder by 4.94 points on average. The same trend is consistently observed across both model sizes and languages. The largest improvement is observed on Retrieval, where the encoder achieves a gain of 10.94 points. These results suggest that, under the same training data budget, encoder models consistently yield better sentence embeddings than decoder models.

\paragraph{Finding 2: Large-scale pre-training substantially improves decoder performance.}
For the Japanese 0.5B models, the adapted pre-trained decoder outperforms the randomly initialized decoder by 5.68 points and surpasses the matched encoder by 0.74 points. The same trend is observed across both model sizes and languages. These results suggest that the success of decoder-based sentence embedding methods is largely attributable to knowledge transferred through large-scale pre-training rather than to the decoder model family itself.

\begin{figure*}[t]
  \centering
  \includegraphics[width=0.93\textwidth]{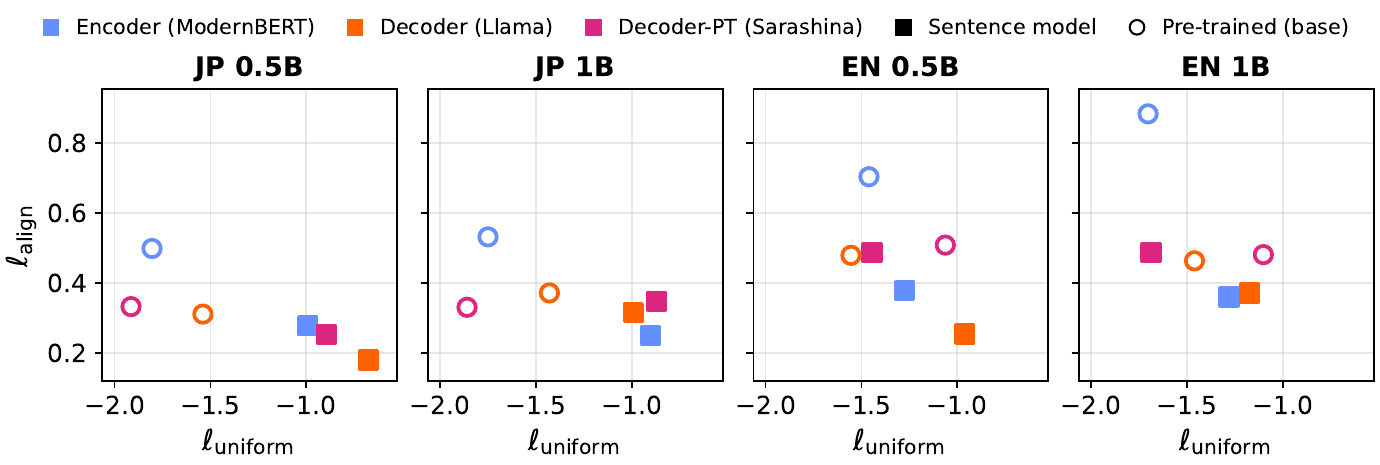}
  \caption{Alignment and uniformity. Squares denote models after sentence embedding training, whereas circles denote the corresponding pre-trained models before sentence embedding training. Decoder circles are measured with bidirectional attention enabled. The alignment loss ($\ell_{\mathrm{align}}$) measures the distance between positive pairs, and the uniformity loss ($\ell_{\mathrm{uniform}}$) measures how uniformly embeddings are distributed on the unit hypersphere. Lower values for both metrics correspond to points closer to the bottom-left corner and indicate better representations.}
  \label{fig:sbench}
\end{figure*}

\subsection{Inference Efficiency}
\begin{table}[t]
  \centering
  \footnotesize
  \setlength{\tabcolsep}{4.5pt}
  \begin{tabular}{@{}lrccccc@{}}
    \toprule
    & & \multicolumn{3}{c}{\textbf{Throughput (texts/s)}} & \multicolumn{2}{c}{\textbf{Memory (MiB)}} \\
    \cmidrule(lr){3-5} \cmidrule(lr){6-7}
     & Len. & Enc. & Dec. & Speedup & Enc. & Dec. \\
    \midrule
    \multirow{2}{*}{JP-0.5B} & 128 & \textbf{641} & 537 & 1.19$\times$ & 1,925 & 1,892 \\
      & 512 & \textbf{362} & 247 & 1.46$\times$ & 3,788 & 3,753 \\
    \midrule
    \multirow{2}{*}{JP-1B} & 128 & \textbf{460} & 369 & 1.25$\times$ & 3,373 & 3,252 \\
      & 512 & \textbf{231} & 162 & 1.43$\times$ & 5,981 & 5,858 \\
    \bottomrule
  \end{tabular}
  \caption{Throughput and peak allocated GPU memory of the Japanese models, measured on an A100 with a batch size of 128 and the decoder's key-value cache disabled. ``Len.'' denotes the sequence length. Speedup is the throughput of the encoder relative to the randomly initialized decoder of the same size. Additional analyses are provided in Appendix~\ref{appendix:efficiency}.}
  \label{tab:efficiency}
\end{table}

Table~\ref{tab:efficiency} shows the measured inference efficiency.

\paragraph{Finding 3: The encoder is faster, whereas memory usage is comparable.}
At a sequence length of 128 tokens, the Japanese 0.5B encoder processes 641 texts/s, compared with 537 texts/s for the matched decoder, corresponding to a 1.19$\times$ speedup. The speedup increases to 1.46$\times$ at 512 tokens. The 1B models exhibit the same trend, with speedups of 1.25$\times$ and 1.43$\times$. In contrast, peak memory usage differs by at most 4\% between the encoder and decoder.
The decoder uses substantially more memory when its key-value cache is enabled: under the same experimental conditions, peak GPU memory usage of the 0.5B decoder increases from 1,892 MiB to 2,852 MiB at a sequence length of 128 tokens. Although these results are specific to the ModernBERT implementation, they highlight the potential throughput advantages of encoder-based sentence embedding models.

\subsection{Representation Quality}

Figure~\ref{fig:sbench} shows the alignment and uniformity analysis across all model configurations.

\paragraph{Finding 4: Alignment and uniformity do not fully explain downstream performance.}
Embedding training generally improves alignment compared with the corresponding pre-trained models. However, for randomly initialized models, uniformity loss increases after embedding training across all settings, e.g., from $-1.80$ to $-0.99$ for the Japanese 0.5B encoder. For the adapted pre-trained decoders, uniformity loss increases in Japanese but decreases in English. Among the embedding-trained models, the matched decoder at 0.5B and the encoder at 1B achieve the best alignment, but neither alignment nor uniformity consistently reflects the relative downstream performance of these models. As shown in Table~\ref{tab:main}, the adapted pre-trained decoder achieves the best downstream performance in every setting, whereas at 0.5B, the matched decoder achieves the best alignment but performs worst among the three models. These results suggest that alignment and uniformity capture only certain aspects of embedding quality. Therefore, these metrics should be interpreted together with downstream evaluation.

\subsection{Language Understanding Ability}

\begin{table}[t]
  \centering
  \small
  \setlength{\tabcolsep}{3.5pt}
  \begin{tabular}{clrrrrr}
    \toprule
    & Model & STS & NLI & CoLA & EQA & CsQA \\
    \midrule
    \multirow{3}{*}{\rotatebox[origin=c]{90}{JP 0.5B}}
    & Encoder & 84.62 & 86.57 & 7.78 & \bfseries 74.58 & 79.54 \\
    & Decoder & 77.38 & 78.35 & \bfseries 10.94 & 55.65 & 62.29 \\
    & Decoder-PT & \bfseries 88.32 & \bfseries 90.96 & 7.78 & 72.95 & \bfseries 91.69 \\
    \midrule
    \multirow{3}{*}{\rotatebox[origin=c]{90}{JP 1B}}
    & Encoder & 85.40 & 87.47 & 8.13 & \bfseries 75.19 & 80.70 \\
    & Decoder & 77.62 & 81.39 & 11.00 & 58.07 & 64.97 \\
    & Decoder-PT & \bfseries 88.30 & \bfseries 92.24 & \bfseries 37.67 & 73.24 & \bfseries 94.46 \\
    \midrule
    \multirow{3}{*}{\rotatebox[origin=c]{90}{EN 0.5B}}
    & Encoder & 82.53 & 80.20 & 40.36 & 85.45 & 40.70 \\
    & Decoder & 67.24 & 73.01 & 0.29 & 63.85 & 27.68 \\
    & Decoder-PT & \bfseries 88.54 & \bfseries 84.91 & \bfseries 47.09 & \bfseries 87.45 & \bfseries 66.58 \\
    \midrule
    \multirow{3}{*}{\rotatebox[origin=c]{90}{EN 1B}}
    & Encoder & 83.95 & 82.09 & 49.34 & 87.13 & 45.05 \\
    & Decoder & 76.22 & 75.39 & 22.29 & 68.25 & 31.45 \\
    & Decoder-PT & \bfseries 90.17 & \bfseries 87.63 & \bfseries 52.86 & \bfseries 88.59 & \bfseries 73.63 \\
    \bottomrule
  \end{tabular}
  \caption{Language understanding performance. STS reports Spearman correlation, NLI reports accuracy, acceptability reports MCC, extractive QA reports F1, and CommonsenseQA reports accuracy. Bold values indicate the best result within each block.}
  \label{tab:diagnostics}
\end{table}

Table~\ref{tab:diagnostics} summarizes the language understanding abilities of each model after language model pre-training and before sentence embedding training. Because these diagnostics are designed to assess the abilities acquired during pre-training, we fine-tune each model using its original attention mechanism. Thus, the decoders retain causal attention. Appendix~\ref{appendix:diag} provides detailed settings. This analysis allows us to examine which language understanding abilities acquired during pre-training contribute to sentence embedding performance.

\paragraph{Finding 5: Sentence-level language understanding follows the same trend as sentence embedding performance.}

On semantic textual similarity, natural language inference, and commonsense reasoning tasks, the randomly initialized encoder outperforms the randomly initialized decoder in every setting. However, large-scale pre-training reverses this ordering. The pre-trained decoder achieves the best performance in all evaluated settings. This ordering exactly matches the sentence embedding results in Table~\ref{tab:main}.
In contrast, extractive question answering follows a different pattern. On JSQuAD and SQuAD, which require token-level positional understanding, the randomly initialized encoder outperforms the randomly initialized decoder across all settings, whereas the pre-trained decoder surpasses the encoder only in English.
These results suggest that sentence-level language understanding acquired during pre-training is more closely associated with sentence embedding quality than token-level span extraction ability.

\subsection{Effect of Architecture and Budget}
\label{sec:decoder-architecture}

To examine whether our findings are robust across decoder architectures and pre-trained models, we additionally evaluate several English 0.5B models. As randomly initialized decoders, we build models with the Mistral configuration \cite{mistral} and the Qwen2.5 configuration \cite{qwen25}. Both models use the same training data and training procedure as the Llama models. As adapted pre-trained decoders, we use Qwen2.5-0.5B and InstructLM-500M \cite{instructlm}, which are based on the Qwen2.5 and Mistral architectures, respectively. Links to all models are provided in Appendix~\ref{appendix:licenses}. Each randomly initialized model shares the architecture and tokenizer of its corresponding pre-trained model. Therefore, this comparison isolates the effect of the underlying decoder model while controlling for all other factors. Table~\ref{tab:arch} reports the results of this additional comparison.

\begin{table}[t]
  \centering
  \small
  \setlength{\tabcolsep}{1.3pt}
  \begin{tabular}{@{}llcccccc@{}}
    \toprule
     & \bfseries Model & \bfseries Avg. & \bfseries Retr. & \bfseries STS & \bfseries Class. & \bfseries Rerank. & \bfseries Clust. \\
    \midrule
    Dec. & Qwen2.5 & 51.73 & 36.70 & 73.86 & 65.20 & 42.00 & 40.90 \\
    Dec. & Mistral & 52.70 & 38.44 & 75.04 & 66.60 & 42.53 & 40.87 \\
    \midrule
    Dec.-PT & Qwen2.5 & 57.19 & 44.18 & 79.48 & 71.92 & 45.23 & \bfseries 45.15 \\
    Dec.-PT & Mistral & \bfseries 57.73 & \bfseries 46.11 & \bfseries 80.01 & \bfseries 73.19 & \bfseries 45.71 & 43.62 \\
    \bottomrule
  \end{tabular}
  \caption{Additional architecture comparison at 0.5B in English on the MTEB subset. The top block shows randomly initialized decoder variants, and the bottom block shows adapted pre-trained models. Each randomly initialized model shares the architecture and tokenizer of its corresponding pre-trained model. Note that ``Dec.-PT Mistral'' denotes the adapted InstructLM-500M model.}
  \label{tab:arch}
\end{table}

\paragraph{Finding 6: The results are consistent across decoder architectures.}
Comparing Tables~\ref{tab:main} and~\ref{tab:arch}, the randomly initialized Mistral and Qwen2.5 decoders remain below the encoder by 0.96 and 1.93 points on average, respectively, and neither configuration surpasses the encoder. In contrast, adapting Qwen2.5-0.5B and InstructLM-500M improves performance by 5.46 and 5.03 points over their randomly initialized counterparts. Randomly initialized decoders never outperform the randomly initialized encoder, whereas large-scale pre-training consistently provides substantial improvements. These results show that the observed ordering is consistent across decoder architectures and pre-trained models, indicating that the conclusions are not specific to a particular decoder implementation.

\begin{table}[t]
  \centering
  \small
  \setlength{\tabcolsep}{2.2pt}
  \begin{tabular}{@{}llcccccc@{}}
    \toprule
     & \bfseries Model & \bfseries Avg. & \bfseries Retr. & \bfseries STS & \bfseries Class. & \bfseries Rerank. & \bfseries Clust. \\
    \midrule
    \multirow{2}{*}{25\%} & Encoder & \bfseries 51.9 & \bfseries 37.0 & \bfseries 74.7 & \bfseries 65.3 & \bfseries 42.2 & \bfseries 40.2 \\
     & Decoder & 49.8 & 34.7 & 71.0 & 64.0 & 41.0 & 38.2 \\
    \midrule
    \multirow{2}{*}{50\%} & Encoder & \bfseries 53.2 & \bfseries 39.5 & \bfseries 75.8 & \bfseries 66.7 & \bfseries 43.2 & \bfseries 40.9 \\
     & Decoder & 51.3 & 36.4 & 73.2 & 65.2 & 41.9 & 39.7 \\
    \midrule
    \multirow{2}{*}{100\%} & Encoder & \bfseries 53.7 & \bfseries 40.3 & \bfseries 76.0 & \bfseries 67.3 & \bfseries 42.9 & \bfseries 41.7 \\
     & Decoder & 52.1 & 37.3 & 74.1 & 66.3 & 42.3 & 40.6 \\
    \bottomrule
  \end{tabular}
  \caption{Effect of the pre-training budget for the English 0.5B models on the MTEB subset. 25\% and 50\% denote checkpoints taken at that fraction of the first pre-training stage, without the second stage, and 100\% denotes the fully pre-trained models in Table~\ref{tab:main}. Bold values indicate the better result at each budget.}
  \label{tab:budget}
\end{table}

\paragraph{Finding 7: The encoder leads at every pre-training budget.}
To examine whether the encoder's lead depends on the pre-training budget, we use intermediate checkpoints of the English 0.5B encoder and decoder at about 25\% and 50\% of the first pre-training stage and apply the same sentence embedding training to them. Table~\ref{tab:budget} shows that the encoder outperforms the decoder across all five task families at every budget, by 2.1, 1.9, and 1.5 points on average at 25\%, 50\%, and 100\%, respectively. Notably, the encoder at 25\% nearly matches the decoder at 100\% (51.9 vs. 52.1). Although the performance gap narrows as the pre-training budget increases, the relative ordering remains unchanged across all evaluated budgets.

\subsection{Analysis of Adaptation Dynamics}
\label{sec:mechanism}

To investigate the training dynamics of sentence embedding models, we run inference-only measurements on checkpoints saved at intervals of 100 to 500 training steps, with no task-specific fine-tuning. This allows us to directly evaluate the sentence embedding quality of each checkpoint without the influence of downstream-task adaptation. As an attention statistic, we measure the future-attention mass, defined as the proportion of attention directed from each query position to future tokens. This value is 0.5 under uniform attention and 0 under causal attention. Embedding quality is tracked by the Spearman correlation on the STS-B validation set using mean pooling. Appendix~\ref{appendix:mechanism} provides detailed measurement settings and additional analyses.

\begin{figure*}[t]
  \centering
  \includegraphics[width=0.99\textwidth]{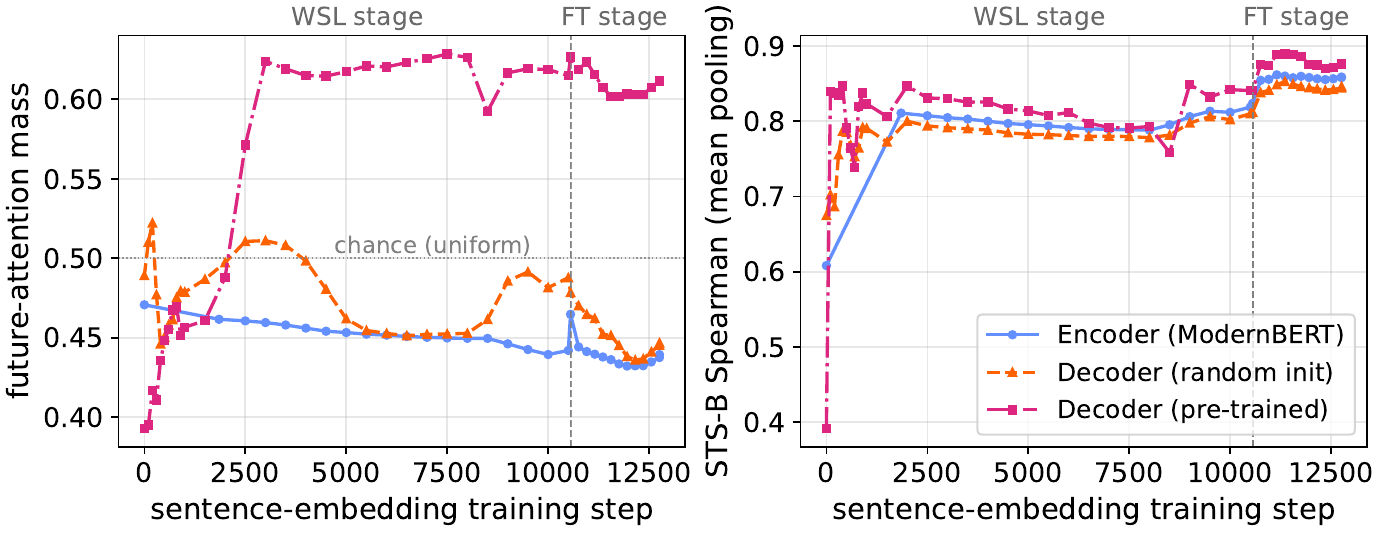}
  \caption{Adaptation trajectories of the English 1B models. Left: future-attention mass averaged across layers, where 0.5 corresponds to uniform attention and 0 to causal attention. The vertical dashed line marks the transition from weakly supervised learning (WSL) to supervised fine-tuning (FT). Step 0 denotes the model before sentence embedding training, with decoder models measured immediately after bidirectionalization. Right: sentence embedding quality measured by the Spearman correlation on the STS-B validation set using mean-pooled embeddings.}
  \label{fig:mechanism}
\end{figure*}

\paragraph{Finding 8: Embedding quality improves before attention patterns stabilize.}

Figure~\ref{fig:mechanism} shows the adaptation trajectories of the English 1B models. The adapted pre-trained decoder rapidly improves on semantic textual similarity, increasing from 0.39 to 0.84 within the first 100 weakly supervised learning steps.
In contrast, the future-attention mass remains at 0.39 at step 100 and reaches 0.62 only at step 3,000. These results suggest that sentence embedding performance improves substantially earlier than the underlying attention patterns stabilize.
Moreover, the trajectories largely converge by the end of the weakly supervised learning stage, with only minor changes observed during the subsequent supervised fine-tuning stage, where STS-B rises by a further 0.03 to 0.04 and the future-attention mass falls by 0.01 to 0.03.
The randomly initialized decoder does not show the same shift toward future tokens. Its future-attention mass remains between 0.44 and 0.52, ending at 0.45, close to the encoder's 0.44. Its final STS-B score is also lower than the encoder's, consistent with the downstream results in Table~\ref{tab:main}. Since we do not incorporate MNTP, the decoder trajectories characterize adaptation directly from pre-trained weights.

\section{Discussions and Recommendations}
\label{sec:discussion}

\paragraph{Pre-training objective.}
Under identical conditions, the randomly initialized encoder consistently outperforms the randomly initialized decoder on downstream sentence embedding tasks. We believe that this gap arises primarily from the pre-training objective, although our experimental design varies both the objective and the block design of each model family (see Section~\ref{sec:setup}). Neither objective operates at the sentence level. MLM predicts individual masked tokens, CLM predicts the next token, and both families obtain sentence embeddings only afterward by pooling token representations and applying contrastive learning. The objectives differ in what each token representation encodes. During pre-training, MLM conditions each token representation on both its left and right contexts, whereas CLM conditions it only on its left context. Consequently, mean pooling after MLM aggregates representations that already incorporate information from the entire sentence. Considering these differences, the encoder's advantage may seem intuitive. However, empirically verifying it requires controlling for the scale of pre-training used in existing decoder-based methods.

\paragraph{Adaptation to sentence embeddings.}
Since all models employ the same contrastive training and use mean pooling during both training and inference (see Section~\ref{sec:training-setting}), the gap does not arise from differences in the adaptation procedure, including the pooling method. Appendix~\ref{appendix:mechanism} provides additional analysis showing that switching the trained English 1B models to last-token pooling at inference degrades the performance of both the encoder and the decoders.
Although LLM2Vec introduces bidirectional attention during sentence embedding training, our training dynamics analysis in Section~\ref{sec:mechanism} suggests that representations acquired through causal language modeling change only gradually during adaptation. Embedding quality improves rapidly, whereas the attention structure continues to reorganize for about 3,000 steps. These observations suggest that simply replacing the causal attention mask with bidirectional attention may be insufficient to fully overcome the representation bias introduced during pre-training. This interpretation is consistent with the downstream results, where the encoder outperforms the randomly initialized decoder across both model sizes and languages.

\paragraph{Pre-trained Knowledge.}

The adapted pre-trained decoder consistently surpasses the matched encoder. For instance, the Japanese 0.5B model exceeds the encoder, with even larger improvements at 1B and in English. A likely explanation is the substantially larger pre-training investment. The randomly initialized models are pre-trained on 10B or 20B tokens, whereas the adapted decoder has been pre-trained on roughly 10T tokens. These results suggest that knowledge acquired during large-scale pre-training transfers effectively to sentence embedding.
However, the benefit of adaptation is not explained solely by pre-training volume. As shown in Section~\ref{sec:decoder-architecture}, adapting InstructLM-500M, pre-trained on roughly 100B tokens, achieves performance comparable to adapting Qwen2.5-0.5B, pre-trained on roughly 18T tokens. These observations suggest that factors beyond pre-training volume, such as corpus quality and composition, also contribute to sentence embedding performance.

\paragraph{Practical guidance.}

Our results provide a consistent guideline across both languages and model sizes. When training sentence embedding models from scratch, encoder models are the preferred choice because they achieve higher downstream performance, and the ModernBERT encoder also encodes text faster with comparable memory use. When a large pre-trained decoder is available, adapting it achieves the best overall performance without requiring additional language model pre-training.
The optimal choice nevertheless depends on the target application. For example, on the Japanese 0.5B Retrieval task, the encoder trained on 10B tokens outperforms the decoder adapted from a model pre-trained on 10T tokens, as shown in Table~\ref{tab:main}. Therefore, when retrieval performance is the primary objective, a compact encoder may provide the best trade-off between effectiveness and computational cost.

\section{Conclusion}

In this study, we investigated a controlled comparison of encoder- and decoder-based sentence embedding models by matching parameter count, tokenizer, training data, and training procedure across Japanese and English at 0.5B and 1B parameters. Under identical conditions, the randomly initialized encoder consistently outperformed the randomly initialized decoder across languages, model sizes, and decoder architectures, within pre-training budgets of up to 20B tokens. In contrast, adapting a large pre-trained decoder achieved the best overall sentence embedding performance, demonstrating that the success of modern decoder-based sentence embedding methods primarily derives from transferred pre-training knowledge rather than from the decoder model family itself.

In addition, our analyses suggest that sentence-level language understanding acquired during pre-training is more closely associated with sentence embedding quality than token-level span extraction ability. Our training dynamics analysis further shows that sentence embedding performance improves much earlier than attention patterns stabilize.
In practice, encoder models are preferable when training from scratch, whereas adapting a large pre-trained decoder is the most effective strategy when one is available.

\section*{Limitations}

\paragraph{Models.}

We evaluate twelve models covering two languages, two model sizes, and three model types: randomly initialized encoders, randomly initialized decoders, and adapted pre-trained decoders. Across these settings, we observe consistent trends. We further evaluate three decoder architectures, namely Llama, Mistral, and Qwen2.5, and obtain the same conclusions. These experiments provide reasonably broad empirical evidence for our research questions and practical guidance for selecting sentence embedding models.
Nevertheless, our study does not cover larger model scales. Evaluating substantially larger models would further strengthen the generality of our conclusions and reveal whether the observed trends continue to hold. However, controlled pre-training beyond 1B parameters requires considerable computational resources, and such models are also less representative of practical sentence embedding systems, where computational efficiency and the availability of suitable pre-trained checkpoints are important considerations. Therefore, we leave comparisons at larger scales to future work.

\paragraph{Coverage.}

We isolate the effects of the model family and pre-training investment under the Chinchilla scaling law by controlling the model size, tokenizer, training data, and training procedure. However, perfectly controlled comparisons between encoder and decoder models remain difficult. Modern transformer architectures differ in many implementation details beyond the pre-training objective, and masked and causal language modeling inherently employ different training objectives and optimization dynamics.
Nevertheless, our conclusions are consistently reproduced across languages, model sizes, and multiple decoder architectures. These results suggest that our main findings are robust to these implementation differences. At the same time, our conclusions should be interpreted within the scope of the configurations evaluated in this work.
Future work includes extending this controlled comparison to additional languages and a broader range of model architectures. Further analyses of architectural components, such as attention mechanisms and MLP designs, may also provide a more detailed understanding of the factors underlying sentence embedding performance. %

\paragraph{Benchmark and Languages.}

We evaluate our models on JMTEB v2, the corresponding English MTEB tasks, and a broad set of language understanding benchmarks. We focus on Japanese and English because these languages provide mature sentence embedding benchmarks, strong pre-trained models, and sufficient resources for constructing controlled bilingual comparisons. To ensure matched comparisons, we construct the English training and evaluation data to correspond closely to the Japanese resources whenever possible.
Although English is the source language in most sentence embedding studies, we do not believe that using Japanese as the source language fundamentally changes the research question addressed in this work. Our objective is not to compare languages, but to isolate the effects of the model family and pre-training under matched experimental conditions.
Extending this framework to additional languages or benchmarks would require similarly aligned training corpora and evaluation datasets, which are currently difficult to obtain. Thus, we restrict our experiments to a bilingual setting, which we believe is sufficient to answer our research questions, while acknowledging that broader validation remains important future work.

\section*{Ethical Considerations}

\paragraph{Licenses and public resources.}

Appendix~\ref{appendix:licenses} shows the licenses and sources of all datasets, pre-trained models, and other public resources used in this study. We verified the applicable licenses and used each resource in accordance with its terms.

\paragraph{Harmful content.}
We trained on large web-derived corpora, which may contain societal biases or other harmful content. Consequently, such biases may also be reflected in the resulting sentence embeddings. However, since our study builds upon established public datasets and existing pre-trained models rather than introducing new training data or annotation, we do not expect it to introduce additional sources of harmful content beyond those already present in the underlying resources.

\paragraph{Use of AI assistants.}
We used AI assistants, such as DeepL, ChatGPT, and Grammarly, for writing support, including grammatical correction, translation, and language polishing. We also used Codex to assist with routine implementation tasks. All research ideas, experimental design, implementation, analyses, interpretations, and the final manuscript originated from the authors. All AI-generated content was reviewed and verified by the authors.

\appendix

\section{Detailed Training Settings}
\label{appendix:training-settings}

\begin{table}[ht]
  \centering
  \small
  \begin{threeparttable}
    \renewcommand{\arraystretch}{0.95}
    \setlength{\tabcolsep}{3pt}
    \begin{tabular}{llcc}
      \toprule
      \textbf{Stage}\tnote{$\dagger$} & \textbf{Item} & \textbf{0.5B} & \textbf{1B} \\
      \midrule
      \multirow{4}{*}{Pre-training 1}
        & LR & $5\times10^{-4}$ & $3\times10^{-4}$ \\
        & BS & \multicolumn{2}{c}{4096} \\
        & Len & \multicolumn{2}{c}{1024} \\
        & Weight decay & \multicolumn{2}{c}{$1\times10^{-5}$} \\
      \midrule
      \multirow{4}{*}{Pre-training 2}
        & LR & $5\times10^{-5}$ & $3\times10^{-5}$ \\
        & BS & \multicolumn{2}{c}{512} \\
        & Len & \multicolumn{2}{c}{8192} \\
        & Weight decay & \multicolumn{2}{c}{$1\times10^{-5}$} \\
      \midrule
      \multirow{5}{*}{WSL}
        & LR & $5\times10^{-5}$ & $3\times10^{-5}$ \\
        & BS & \multicolumn{2}{c}{4096} \\
        & Len & \multicolumn{2}{c}{256} \\
        & $\tau$ & \multicolumn{2}{c}{0.01} \\
        & Weight decay & \multicolumn{2}{c}{$1\times10^{-5}$} \\
      \midrule
      \multirow{5}{*}{FT}
        & LR & $5\times10^{-6}$ & $3\times10^{-6}$ \\
        & BS & \multicolumn{2}{c}{512} \\
        & Len & \multicolumn{2}{c}{512} \\
        & $\tau$ & \multicolumn{2}{c}{0.01} \\
        & Weight decay & \multicolumn{2}{c}{$1\times10^{-5}$} \\
      \bottomrule
    \end{tabular}
  \end{threeparttable}
  \caption{Training hyperparameters for language model pre-training, weakly supervised learning (WSL), and supervised fine-tuning (FT). Hyperparameters are shared across architectures within each model size and differ across languages only in the training corpora and pre-training token budget. BS denotes the total batch size, Len the maximum input sequence length, and $\tau$ the InfoNCE temperature. The datasets used at each stage are described in Section~\ref{sec:datasets} and this appendix. For Japanese WSL, we sample at most 2M pairs from each subset of \texttt{ruri-dataset-v2-pt}.}
  \label{tab:pretrain-settings}
\end{table}

Table~\ref{tab:pretrain-settings} summarizes the training settings for each stage, namely language model pre-training, weakly supervised learning (WSL), and supervised fine-tuning (FT), for every model size and language. Learning rates are adjusted according to model size, whereas batch sizes are shared across model sizes; both are shared across languages and architectures within the same size. Only the training corpora and pre-training token budgets differ across languages.
Sentence embedding training uses the InfoNCE loss \cite{infonce}. During WSL and FT, we adopt task-homogeneous batching following GTE \cite{gte}, where each batch contains samples from a single dataset. This strategy provides more informative in-batch negatives while reducing padding overhead. All stages use a constant learning rate without warmup, gradient clipping with a maximum norm of 1.0, bfloat16 precision, and FlashAttention-2 \cite{flashattention2}. Both WSL and FT are trained for one epoch.
Table~\ref{tab:arch-config} summarizes the architecture configurations used in the main comparison.

\begin{table}[t]
  \centering
  \small
  \setlength{\tabcolsep}{4pt}
  \begin{tabular}{lcccc}
    \toprule
    & \multicolumn{2}{c}{\textbf{0.5B}} & \multicolumn{2}{c}{\textbf{1B}} \\
    \cmidrule(lr){2-3} \cmidrule(lr){4-5}
    \textbf{Item} & Enc & Dec & Enc & Dec \\
    \midrule
    Layers & 23 & 24 & 24 & 24 \\
    Hidden dim & 1280 & 1280 & 1792 & 1792 \\
    FFN dim & 4480 & 4480 & 6272 & 6272 \\
    Attention heads & 16 & 16 & 16 & 16 \\
    Vocabulary & 102{,}400 & 102{,}400 & 102{,}400 & 102{,}400 \\
    \bottomrule
  \end{tabular}
  \caption{Architecture configurations of the main-comparison models. The decoder uses the Sarashina2.2 architecture with random initialization, whereas the encoder follows the ModernBERT architecture with the number of layers adjusted to match the decoder's non-embedding parameter count. The 0.5B encoder has 23 layers rather than 24 because we keep its hidden and FFN dimensions identical to those of the decoder and prioritize matching the non-embedding parameter count over matching the number of layers. Decoder models use grouped-query attention \cite{gqa} with 8 key-value heads. Encoder models use global attention every three layers and local attention with a window size of 128 in the remaining layers. All models share a vocabulary size of 102{,}400 and a maximum sequence length of 8192.}
  \label{tab:arch-config}
\end{table}

The Japanese sentence embedding training data consists of about 43M WSL pairs sampled from the 34 subsets of ruri-dataset-v2-pt, with at most 2M pairs per subset, and about 630K FT examples from ruri-v3-dataset-ft.
The English sentence embedding training data is constructed from embedding-training-data to match the supervision types used in the Japanese Ruri datasets. WSL uses about 44M sentence pairs from the following 25 subsets:

\begin{description}[leftmargin=*, itemsep=0em, topsep=0.5em, left=0.15em]
  \item[QA:] NQ, SQuAD, TriviaQA, GooAQ, PAQ (capped at 4.6M), Amazon-QA, ELI5, SearchQA, and Yahoo Answers (question--answer, title--answer, and title--question)
  \item[Duplicate questions:] Quora, WikiAnswers, and StackExchange (title--title, body--body, and title--body)
  \item[News and articles:] AGNews, CC-News, NPR, CNN/DailyMail, and XSum
  \item[Paraphrase and summarization:] AltLex, SimpleWiki, Sentence Compression, and WikiHow
\end{description}
All datasets use the versions provided in embedding-training-data. FT consists of triplets from AllNLI, MS MARCO, and Quora duplicate questions together with QA pairs from NQ \cite{nq}, SQuAD \cite{squad}, and TriviaQA \cite{triviaqa} augmented with random negatives, resulting in 1.1M training examples.
Following the Chinchilla scaling law, 1B models are pre-trained on 20B tokens, while 0.5B models are pre-trained on 10B tokens. The English models use matched token budgets. Each model in the additional architecture comparison uses the same 0.5B training settings as Table~\ref{tab:pretrain-settings}; only the tokenizer and decoder architecture differ.

\section{Information of Models and Datasets}
\label{appendix:licenses}

Table~\ref{tab:licenses} summarizes the external models and datasets used in this study, including their distribution pages and license information.

\begin{table*}[t]
  \centering
  \footnotesize
   \setlength{\tabcolsep}{0.5pt}
  \begin{tabular}{@{}lll@{}}
    \toprule
    \bfseries Name & \bfseries URL & \bfseries License \\
    \midrule
    \multicolumn{3}{@{}l}{\itshape Models} \\
    sarashina2.2-0.5b & \href{https://hf.co/sbintuitions/sarashina2.2-0.5b}{\nolinkurl{hf.co/sbintuitions/sarashina2.2-0.5b}} & MIT \\
    sarashina2.2-1b & \href{https://hf.co/sbintuitions/sarashina2.2-1b}{\nolinkurl{hf.co/sbintuitions/sarashina2.2-1b}} & MIT \\
    Qwen2.5-0.5B & \href{https://hf.co/Qwen/Qwen2.5-0.5B}{\nolinkurl{hf.co/Qwen/Qwen2.5-0.5B}} & Apache-2.0 \\
    InstructLM-500M & \href{https://hf.co/instruction-pretrain/InstructLM-500M}{\nolinkurl{hf.co/instruction-pretrain/InstructLM-500M}} & Apache-2.0 \\
    \addlinespace
    \midrule
    \multicolumn{3}{@{}l}{\itshape Datasets} \\
    FineWeb-Edu & \href{https://hf.co/datasets/HuggingFaceFW/fineweb-edu}{\nolinkurl{hf.co/datasets/HuggingFaceFW/fineweb-edu}} & ODC-BY \\
    FineWeb-2-Edu-Japanese & \href{https://hf.co/datasets/hotchpotch/fineweb-2-edu-japanese}{\nolinkurl{hf.co/datasets/hotchpotch/fineweb-2-edu-japanese}} & ODC-BY \\
    wiki40b\_ja & \href{https://hf.co/datasets/fujiki/wiki40b_ja}{\nolinkurl{hf.co/datasets/fujiki/wiki40b\_ja}} & CC BY-SA 4.0 \\
    Wiki-40B (en) & \href{https://hf.co/datasets/google/wiki40b}{\nolinkurl{hf.co/datasets/google/wiki40b}} & CC BY-SA \\
    ruri-dataset-v2-pt & \href{https://hf.co/datasets/cl-nagoya/ruri-dataset-v2-pt}{\nolinkurl{hf.co/datasets/cl-nagoya/ruri-dataset-v2-pt}} & Per-source \\
    ruri-v3-dataset-ft & \href{https://hf.co/datasets/cl-nagoya/ruri-v3-dataset-ft}{\nolinkurl{hf.co/datasets/cl-nagoya/ruri-v3-dataset-ft}} & Per-source \\
    embedding-training-data & \href{https://hf.co/datasets/sentence-transformers/embedding-training-data}{\nolinkurl{hf.co/datasets/sentence-transformers/embedding-training-data}} & Per-source \\
    AllNLI & \href{https://hf.co/datasets/sentence-transformers/all-nli}{\nolinkurl{hf.co/datasets/sentence-transformers/all-nli}} & Per-source \\
    Wikipedia & \href{https://hf.co/datasets/wikimedia/wikipedia}{\nolinkurl{hf.co/datasets/wikimedia/wikipedia}} & CC BY-SA 3.0/GFDL \\
    MIRACL & \href{https://hf.co/datasets/miracl/miracl}{\nolinkurl{hf.co/datasets/miracl/miracl}} & Apache-2.0 \\
    Quora Question Pairs & \href{https://quoradata.quora.com/First-Quora-Dataset-Release-Question-Pairs}{\nolinkurl{quoradata.quora.com/First-Quora-Dataset-Release-Question-Pairs}} & Quora ToS \\
    \bottomrule
  \end{tabular}
  \caption{Links and licenses of the external models and datasets used in this study. Licenses are as stated on each distribution page at the time of writing; entries marked per-source aggregate multiple datasets whose licenses follow the original sources, and Wiki-40B inherits the CC BY-SA terms of its Wikipedia source. The Quora data is subject to Quora's Terms of Service (\url{https://www.quora.com/about/tos}; non-commercial use).}
  \label{tab:licenses}
\end{table*}

\section{English MTEB Subset}
\label{appendix:mteb-tasks}

English models are evaluated on the 37 English MTEB v2 tasks corresponding to the five task families of JMTEB \cite{mteb, mmteb}. For multilingual tasks, we use only the official English subsets, namely STS17 (en-en) and STS22.v2 (en). All scores are computed on the test split, and AmazonCounterfactualClassification uses its \texttt{en} subset. Family scores are computed as the mean over the tasks within each family, and the reported overall score is the macro-average across the five task families.

{\small\raggedright\hyphenpenalty=10000\exhyphenpenalty=10000
\begin{description}[leftmargin=*, itemsep=0.1em, topsep=0.5em, left=0.15em]
  \item[Retrieval (10):] ArguAna, ClimateFEVERHardNegatives, CQADupstackGamingRetrieval, CQADupstackUnixRetrieval, FEVERHardNegatives, FiQA2018, HotpotQAHardNegatives, SCIDOCS, TRECCOVID, Touche2020Retrieval.v3
  \item[STS (9):] BIOSSES, SICK-R, STS12, STS13, STS14, STS15, STS17, STS22.v2, STSBenchmark
  \item[Classification (8):] AmazonCounterfactualClassification, Banking77Classification, ImdbClassification, MTOPDomainClassification, MassiveIntentClassification, MassiveScenarioClassification, ToxicConversationsClassification, TweetSentimentExtractionClassification
  \item[Reranking (2):] AskUbuntuDupQuestions, MindSmallReranking
  \item[Clustering (8):] ArXivHierarchicalClusteringP2P, ArXivHierarchicalClusteringS2S, BiorxivClusteringP2P.v2, MedrxivClusteringP2P.v2, MedrxivClusteringS2S.v2, StackExchangeClustering.v2, StackExchangeClusteringP2P.v2, TwentyNewsgroupsClustering.v2
\end{description}
}

\section{Detailed Efficiency Results}
\label{appendix:efficiency}

\subsection{Measurement Conditions}

This appendix reports two sets of measurements. Both use sentence-transformers \cite{sbert} in bfloat16 precision.

The first set compares the Japanese models on an A100 (80GB) with the decoder's key-value cache disabled (see Tables~\ref{tab:efficiency} and~\ref{tab:efficiency-nocache}). We report throughput and the peak GPU memory allocated by PyTorch at a batch size of 128. Sequence lengths of 128 and 512 use 20K texts sampled from Wiki40B Japanese, and lengths of 1,024 and 2,048 use 4,000 fixed-length inputs.

The second set covers the same models on an A100 (40GB) under the default setting, which keeps the decoder's key-value cache enabled (see Table~\ref{tab:efficiency-default} and Figures~\ref{fig:throughput} to~\ref{fig:memory-extra}). It encodes 100K texts sampled from Wiki40B Japanese at sequence lengths of 128, 256, and 512 with batch sizes from 16 to 256, and reports the peak GPU memory usage of the process.

Since ModernBERT applies local attention with a window of 128 tokens in two of every three layers, these layers do not attend to the full input when the sequence length exceeds the attention window. This implementation characteristic favors the encoder in comparisons with decoder models and should be considered when interpreting the results.

\subsection{Memory and the Key-Value Cache}

\begin{table}[t]
  \centering
  \footnotesize
  \setlength{\tabcolsep}{2pt}
  \begin{tabular}{@{}lrccccc@{}}
    \toprule
    & & \multicolumn{3}{c}{\textbf{Throughput (texts/s)}} & \multicolumn{2}{c}{\textbf{Memory (MiB)}} \\
    \cmidrule(lr){3-5} \cmidrule(lr){6-7}
     & Len. & Enc. & Dec. & Speedup & Enc. & Dec. \\
    \midrule
    \multirow{4}{*}{JP-0.5B} & 128 & 641 & 537 & 1.19$\times$ & 1,925 & 1,892 \\
      & 512 & 362 & 247 & 1.46$\times$ & 3,788 & 3,753 \\
      & 1,024 & 101 & 80 & 1.27$\times$ & 6,273 & 6,235 \\
      & 2,048 & 53 & 39 & 1.34$\times$ & 11,244 & 11,200 \\
    \midrule
    \multirow{4}{*}{JP-1B} & 128 & 460 & 369 & 1.25$\times$ & 3,373 & 3,252 \\
      & 512 & 231 & 162 & 1.43$\times$ & 5,981 & 5,858 \\
      & 1,024 & 63 & 52 & 1.20$\times$ & 9,459 & 9,332 \\
      & 2,048 & 32 & 26 & 1.24$\times$ & 16,414 & 16,281 \\
    \bottomrule
  \end{tabular}
  \caption{Throughput and peak allocated GPU memory of the Japanese models with the decoder's key-value cache disabled, at a batch size of 128. Speedup is computed from unrounded throughput.}
  \label{tab:efficiency-nocache}
\end{table}

Table~\ref{tab:efficiency-nocache} extends Table~\ref{tab:efficiency} to sequence lengths of up to 2,048 tokens. Once the cache is disabled, the peak memory of the decoder stays within 4\% of that of the encoder at every length, whereas the encoder remains 1.19 to 1.46$\times$ faster. The speedup does not grow further at 1,024 and 2,048 tokens, which use fixed-length inputs. In the same setup with the cache enabled, the Japanese 0.5B decoder allocates 2,852 MiB instead of 1,892 MiB at a sequence length of 128. The cache stores keys and values for incremental decoding, which embedding inference never uses. The model weights occupy nearly the same memory, 1,295 MiB for the 0.5B encoder and 1,263 MiB for the 0.5B decoder. The memory gaps in Figures~\ref{fig:memory} and~\ref{fig:memory-extra} are consistent with the cache.

The measurements in Table~\ref{tab:efficiency} pad each batch to its longest input, which makes about 5\% of the tokens padding at a sequence length of 128, so unpadding in ModernBERT skips at most about 5\% of the input tokens. When we instead pad every input to 128 tokens (about 12\% padding) in a separate run with the cache enabled, the throughput of the encoder is nearly unchanged (640 to 634 texts/s), whereas that of the decoder falls from 510 to 406 texts/s. In this run, the encoder's advantage is larger with more padding.

\subsection{Throughput and Memory Usage under the Default Setting}

\begin{figure}[!t]
  \centering
  \includegraphics[width=\linewidth]{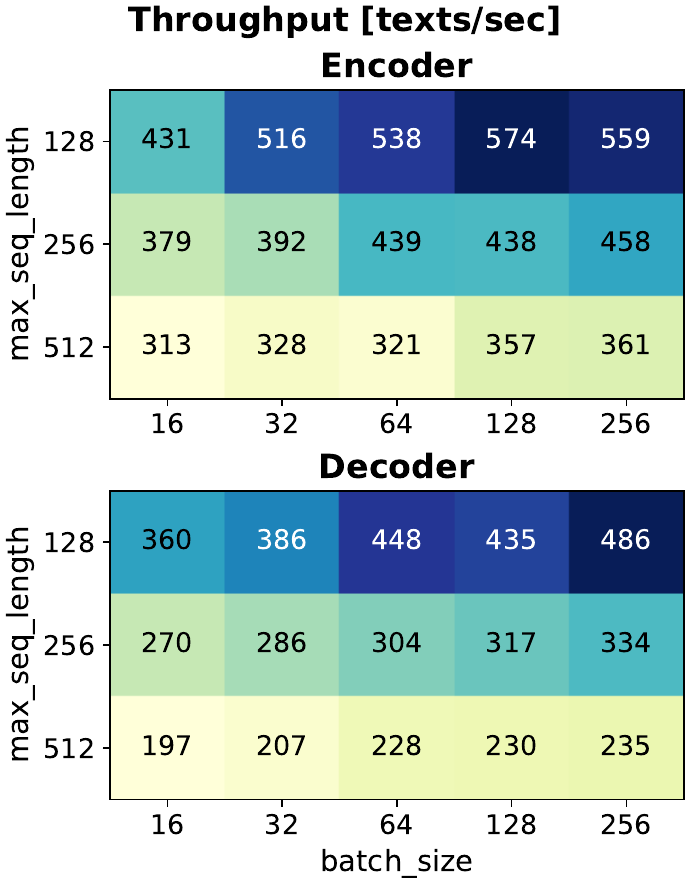}
  \caption{Throughput of the Japanese 0.5B models across batch sizes and sequence lengths under the default setting (key-value cache enabled). Top: encoder; bottom: decoder.}
  \label{fig:throughput}
\end{figure}

\begin{figure}[!t]
  \centering
  \includegraphics[width=\linewidth]{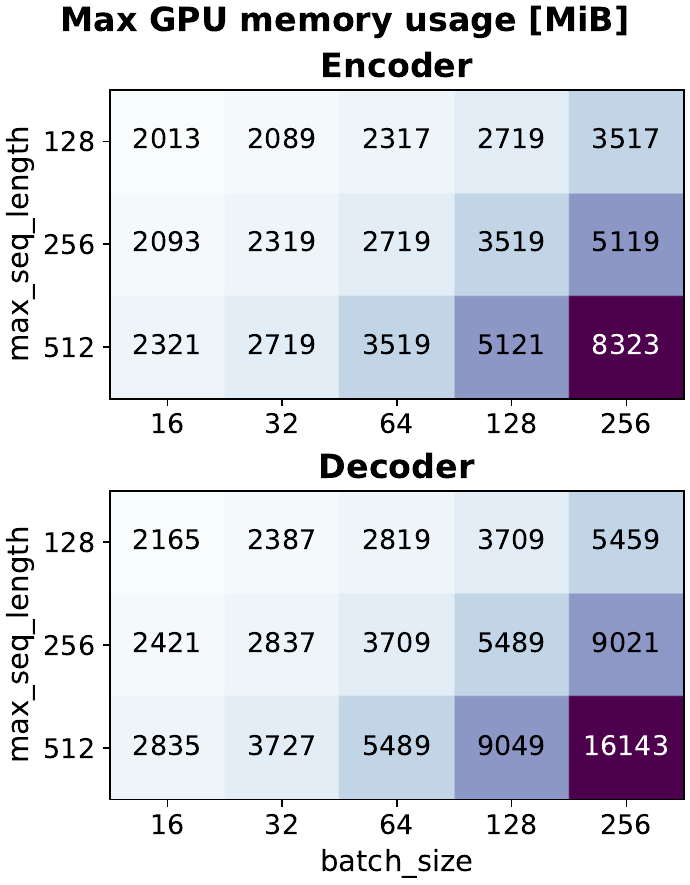}
  \caption{GPU memory usage of the Japanese 0.5B models across batch sizes and sequence lengths under the default setting (key-value cache enabled). Top: encoder; bottom: decoder.}
  \label{fig:memory}
\end{figure}

Figures~\ref{fig:throughput} and \ref{fig:memory} show throughput and GPU memory usage across different batch sizes and sequence lengths for the Japanese 0.5B models under the default setting.
The encoder achieves higher throughput than the matched decoder under every measured condition. The decoder uses more GPU memory, and the difference grows with the sequence length because the key-value cache grows with it. Table~\ref{tab:efficiency-default} lists the values of both model sizes at a sequence length of 128 and a batch size of 128.

\begin{table}[t]
  \centering
  \footnotesize
  \setlength{\tabcolsep}{2pt}
  \begin{tabular}{@{}lccccc@{}}
    \toprule
    & \multicolumn{3}{c}{\textbf{Throughput (texts/s)}} & \multicolumn{2}{c}{\textbf{Memory (MiB)}} \\
    \cmidrule(lr){2-4} \cmidrule(lr){5-6}
     & Enc. & Dec. & Speedup & Enc. & Dec. \\
    \midrule
    JP-0.5B & 574 & 435 & 1.32$\times$ & 2,719 & 3,709 \\
    JP-1B   & 436 & 354 & 1.23$\times$ & 4,239 & 5,497 \\
    \bottomrule
  \end{tabular}
  \caption{Throughput and peak GPU memory usage under the default setting (key-value cache enabled), measured on an A100 (40GB) with a sequence length of 128 and a batch size of 128.}
  \label{tab:efficiency-default}
\end{table}

\subsection{Results for the 1B Models}
\begin{figure}[t]
  \centering
  \includegraphics[width=\linewidth]{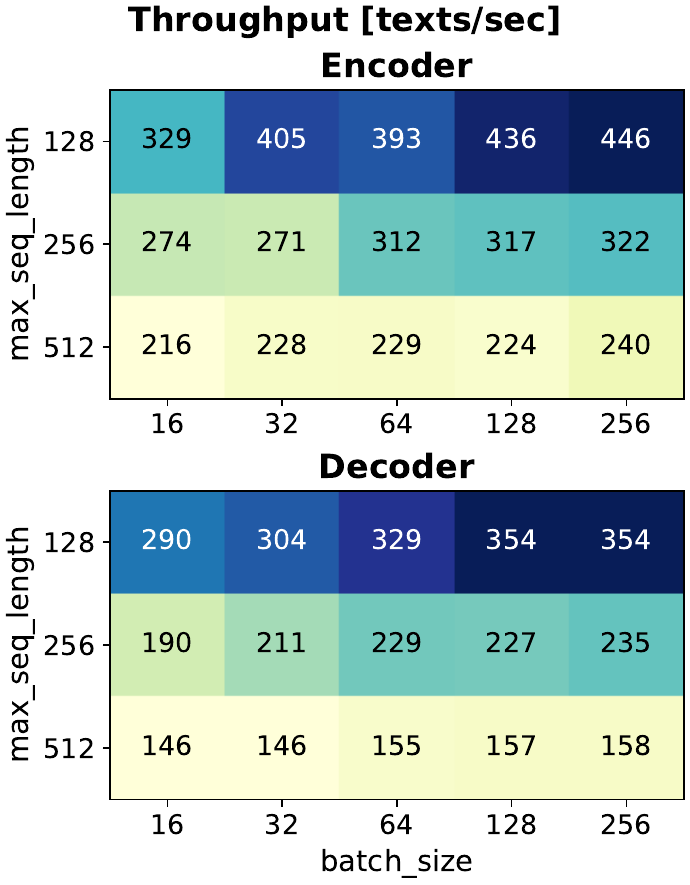}
  \caption{Throughput of the Japanese 1B models across batch sizes and sequence lengths under the default setting (key-value cache enabled). Top: encoder; bottom: decoder.}
  \label{fig:throughput-extra}
\end{figure}

\begin{figure}[t]
  \centering
  \includegraphics[width=\linewidth]{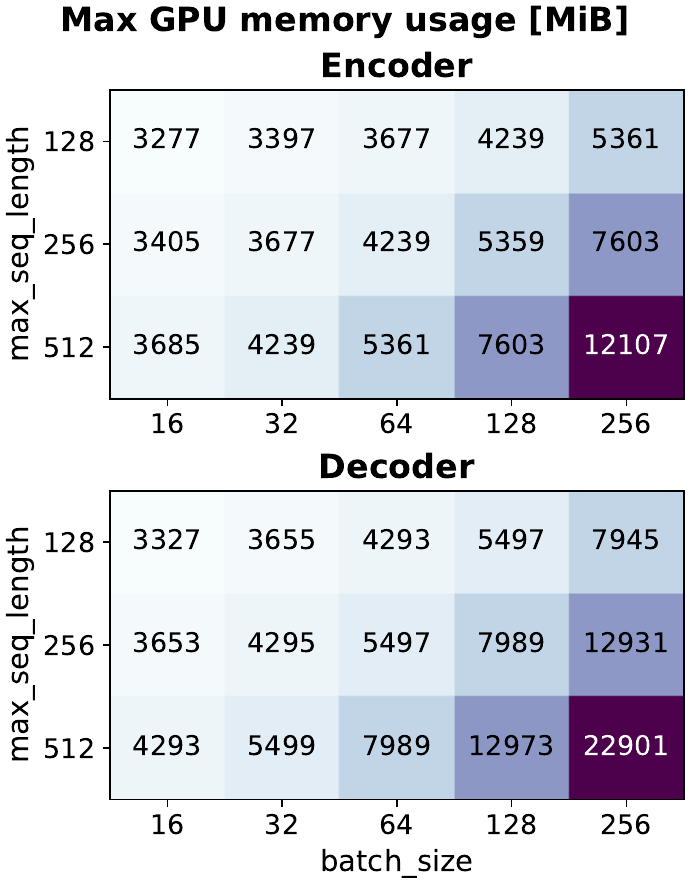}
  \caption{GPU memory usage of the Japanese 1B models across batch sizes and sequence lengths under the default setting (key-value cache enabled). Top: encoder; bottom: decoder.}
  \label{fig:memory-extra}
\end{figure}

Figures~\ref{fig:throughput-extra} and \ref{fig:memory-extra} present the corresponding results for the Japanese 1B models. The encoder again achieves higher throughput under every measured condition, and the memory gaps are again consistent with the key-value cache.

\section{Settings: Language Understanding}
\label{appendix:diag}

\begin{table*}[t]
  \centering
  \small
  \begin{tabular}{clccccc}
    \toprule
    & \textbf{Model} & \makecell{\textbf{JSTS}\\ \footnotesize Pearson/Spearman} & \textbf{JNLI} & \makecell{\textbf{JCoLA}\\ \footnotesize Acc/MCC} & \makecell{\textbf{JSQuAD}\\ \footnotesize EM/F1} & \textbf{JCommonsenseQA} \\
    \midrule
    & Chance & 0.00/0.00 & 55.46 & 83.93/0.00 & 0.00/0.00 & 20.00 \\
    \midrule
    \multirow{3}{*}{\rotatebox[origin=c]{90}{JP 0.5B}}
    & Encoder & 88.71/84.62 & 86.57 & 84.05/7.78 & \textbf{74.52}/\textbf{74.58} & 79.54 \\
    & Decoder & 82.66/77.38 & 78.35 & \textbf{84.16}/\textbf{10.94} & 55.07/55.65 & 62.29 \\
    & Decoder-PT & \textbf{91.89}/\textbf{88.32} & \textbf{90.96} & 84.05/7.78 & 72.67/72.95 & \textbf{91.69} \\
    \midrule
    \multirow{3}{*}{\rotatebox[origin=c]{90}{JP 1B}}
    & Encoder & 89.37/85.40 & 87.47 & 84.05/8.13 & \textbf{75.06}/\textbf{75.19} & 80.70 \\
    & Decoder & 83.37/77.62 & 81.39 & 84.16/11.00 & 57.54/58.07 & 64.97 \\
    & Decoder-PT & \textbf{92.10}/\textbf{88.30} & \textbf{92.24} & \textbf{85.32}/\textbf{37.67} & 72.92/73.24 & \textbf{94.46} \\
    \bottomrule
  \end{tabular}
  \caption{Language-understanding diagnostics of the Japanese pre-trained models. Rows correspond to the Encoder, Decoder, and Decoder-PT models in Table~\ref{tab:main}. Following \citet{sakai2024templates}, JSTS is evaluated by Pearson and Spearman correlation, JNLI and JCommonsenseQA by accuracy, JCoLA by accuracy and Matthews correlation, and JSQuAD by exact match and F1. All scores are reported as percentages. Chance denotes the no-learning baseline, computed as the majority-class accuracy for classification tasks, the uniform-choice accuracy for multiple-choice tasks, and zero otherwise. Bold values indicate the best result within each model size. The identical JCoLA values of the Japanese 0.5B Encoder and Decoder-PT are independent measurements.
}
  \label{tab:jglue}
\end{table*}

\begin{table*}[t]
  \centering
  \small
  \begin{tabular}{lccccc}
    \toprule
    \textbf{Model} & \makecell{\textbf{STS-B}\\ \footnotesize Pearson/Spearman} & \textbf{MNLI} & \makecell{\textbf{CoLA}\\ \footnotesize Acc/MCC} & \makecell{\textbf{SQuAD}\\ \footnotesize EM/F1} & \textbf{CommonsenseQA} \\
    \midrule
    Chance & 0.00/0.00 & 35.45 & 69.13/0.00 & 0.00/0.00 & 20.00 \\
    \midrule
    ModernBERT-EN-0.5B & 82.77/82.53 & 80.20 & 76.61/40.36 & 77.56/85.45 & 40.70 \\
    Llama-EN-0.5B & 67.67/67.24 & 73.01 & 69.03/0.29 & 51.32/63.85 & 27.68 \\
    Mistral-EN-0.5B & 75.00/75.16 & 74.20 & 69.80/21.01 & 53.26/65.70 & 22.60 \\
    Qwen2.5-Scratch-EN-0.5B & 64.63/64.84 & 74.51 & 68.94/$-$2.93 & 50.27/62.68 & 28.75 \\
    Sarashina-0.5B & \textbf{88.31}/\textbf{88.54} & \textbf{84.91} & \textbf{78.14}/\textbf{47.09} & \textbf{79.08}/\textbf{87.45} & \textbf{66.58} \\
    \midrule
    ModernBERT-EN-1B & 84.09/83.95 & 82.09 & 79.29/49.34 & 79.47/87.13 & 45.05 \\
    Llama-EN-1B & 76.33/76.22 & 75.39 & 71.05/22.29 & 55.88/68.25 & 31.45 \\
    Sarashina-1B & \textbf{90.17}/\textbf{90.17} & \textbf{87.63} & \textbf{80.54}/\textbf{52.86} & \textbf{81.22}/\textbf{88.59} & \textbf{73.63} \\
    \bottomrule
  \end{tabular}
  \caption{Language-understanding diagnostics of the English pre-trained models. Each task corresponds to its Japanese counterpart in Table~\ref{tab:jglue}. Following the same evaluation protocol, STS-B is evaluated by Pearson and Spearman correlation, MNLI and CommonsenseQA by accuracy, CoLA by accuracy and Matthews correlation, and SQuAD by exact match and F1. All scores are reported as percentages. Chance denotes the no-learning baseline. Bold values indicate the best result within each model size.}
  \label{tab:glue}
\end{table*}

The diagnostics reported in the main text fine-tune each pre-trained model using task-specific heads from Hugging Face Transformers \cite{transformerslib}. Classification and regression tasks (JSTS, JNLI, JCoLA, STS-B, MNLI, and CoLA) use a sequence-classification head. Extractive question answering (JSQuAD and SQuAD) uses a span-prediction head. Multiple-choice tasks (JCommonsenseQA and CommonsenseQA) use one logit from a sequence-classification head for each candidate answer and are optimized with softmax cross-entropy. The classification head reads the first token for encoder models and the last non-padding token for decoder models.

Decoder models are fine-tuned with the same causal attention used during pre-training; no bidirectionalization is applied. The diagnostics are meant to measure what each model acquires during pre-training, before any sentence embedding training, so each model keeps the attention it was pre-trained with. For all models except Qwen2.5, the tokenizer is configured to prepend a beginning-of-sequence (BOS) token to every input so that causal attention has a dedicated first position. The Qwen2.5 tokenizer does not define a BOS token, and its inputs are therefore left unchanged. For extractive question answering, training windows that do not contain the answer span are discarded.

Training runs for three epochs with a learning rate of $2\times10^{-5}$ for 0.5B models and $1\times10^{-5}$ for 1B models, an effective batch size of 32, and maximum sequence lengths of 512 for classification and regression, 384 for extractive question answering, and 128 for multiple-choice tasks. All other optimization settings follow Appendix~\ref{appendix:training-settings}. We evaluate the checkpoint with the best validation performance. All experiments are conducted on a single NVIDIA A100 GPU.

\section{Details of the Mechanistic Analysis}
\label{appendix:mechanism}

We describe the protocol used for the adaptation-dynamics analysis in Section~\ref{sec:mechanism} and present the complete results underlying the findings.

\paragraph{Measurement conditions.}
The mechanistic analysis runs inference only using an attention implementation that exposes attention weights. Bidirectionalized decoders receive an explicit bidirectional attention mask so that they behave exactly as during training. We verify this implementation with a unit test. Measurement sentences come from the same sources as the representation-quality evaluation. We use 2,000 MIRACL positive pairs for Japanese and 2,000 NLI premise--hypothesis pairs for English. Sentence embedding quality is evaluated on the JSTS and STS-B validation sets. Attention statistics are computed over 32 sentences with a maximum sequence length of 128 tokens. For the English 1B trajectories, we measure checkpoints every 100 steps up to step 1,000 of weakly supervised learning, every 500 steps afterwards, and every 200 steps during supervised fine-tuning, in addition to the last checkpoint of each stage. For the encoder, the retained checkpoints of weakly supervised learning are step 1,850 and every 500 steps from step 2,500.
To evaluate the robustness of the measurements with respect to the choice of evaluation sentences, we estimate 95\% confidence intervals of the mean using sentence-level bootstrap over 256 held-out Wikipedia sentences. For the final English 1B models, the future-attention mass is 0.491 with a 95\% confidence interval of [0.484, 0.497] for the adapted pre-trained decoder, 0.413 with [0.411, 0.416] for the randomly initialized decoder, and 0.430 with [0.428, 0.432] for the encoder. The non-overlapping confidence intervals indicate that the observed ordering on these sentences is robust to sentence sampling. The absolute values vary with the measurement corpus. For example, the adapted pre-trained decoder reaches 0.61 on the NLI sentences used in Figure~\ref{fig:mechanism} and 0.49 on held-out Wikipedia sentences. The adapted pre-trained decoder has the largest value on both sentence sets, but the encoder and the randomly initialized decoder swap places, with 0.44 and 0.45 on the NLI sentences. Therefore, we compare values only within the same sentence set and use the NLI-based measurements as the primary analysis.

\paragraph{Non-monotonicity in pre-training volume.}
We further compare adaptation sources with different pre-training volumes for the English 0.5B models. The randomly initialized Llama, Mistral, and Qwen2.5 models are pre-trained on 10B tokens, whereas InstructLM is pre-trained on 100B tokens, Sarashina on approximately 10T tokens, and Qwen2.5 on approximately 18T tokens. After sentence embedding training, STS improves only mildly from 0.84 to 0.88. In contrast, the future-attention mass is non-monotonic, taking values of 0.36 to 0.42 for the 10B models, 0.22 for InstructLM, 0.47 for Sarashina, and 0.52 for Qwen2.5. The effective rank of the embedding spectrum \cite{garrido2023rankme} is likewise non-monotonic, with values of 37 to 40 for the 10B models, 41.6 for InstructLM, 40.9 for Sarashina, and 42.4 for Qwen2.5. These observations are consistent with the non-monotonic adaptation benefit observed in downstream performance, as discussed in Section~\ref{sec:discussion}.

\paragraph{Pooling.}
Table~\ref{tab:pooling} compares mean pooling, which all models use during training, with last-token pooling applied only at inference, for the English 1B models. Switching to last-token pooling lowers the average by 25.9 points for the encoder, 23.2 points for the decoder, and 23.3 points for the adapted pre-trained decoder, and STS and Retrieval degrade the most. Because the encoder drops at least as much as the decoders, the drop reflects the mismatch with the pooling used in training.

\begin{table}[t]
  \centering
  \small
  \setlength{\tabcolsep}{1.6pt}
  \begin{tabular}{@{}llcccccc@{}}
    \toprule
    \bfseries Model & \bfseries Pool. & \bfseries Avg. & \bfseries Retr. & \bfseries STS & \bfseries Class. & \bfseries Rerank. & \bfseries Clust. \\
    \midrule
    \multirow{2}{*}{Encoder} & mean & 54.8 & 42.3 & 76.7 & 68.5 & 43.9 & 42.5 \\
     & last & 28.9 & 7.4 & 30.9 & 44.2 & 35.7 & 26.3 \\
    \midrule
    \multirow{2}{*}{Decoder} & mean & 52.9 & 39.0 & 75.5 & 66.4 & 42.5 & 41.3 \\
     & last & 29.7 & 5.1 & 34.0 & 46.8 & 34.1 & 28.8 \\
    \midrule
    \multirow{2}{*}{Decoder-PT} & mean & 57.6 & 44.9 & 79.8 & 73.5 & 46.0 & 43.6 \\
     & last & 34.3 & 19.6 & 36.0 & 52.9 & 36.2 & 27.0 \\
    \bottomrule
  \end{tabular}
  \caption{Effect of the pooling method at inference for the English 1B models on the MTEB subset. All models are trained with mean pooling. ``last'' applies last-token pooling at inference only.}
  \label{tab:pooling}
\end{table}

\paragraph{Other observations.}
We find no evidence that a few high-variance dimensions dominate the embeddings of the twelve trained models in Table~\ref{tab:main} \cite{timkey2021rogue,rudman2023outlier}; the largest dimension accounts for less than 0.5\% of the total variance, and ablating the top eight variance dimensions changes STS by at most 0.002. The share of last-layer attention on the first token, where decoder models are reported to concentrate attention \cite{xiao2024efficient}, is at most 7\% for the decoders among these models and changes in no consistent direction through adaptation.

\section{Computational Resources}
\label{appendix:compute}

Across all experiments reported in this paper, we used a total of approximately 13{,}300 GPU hours, covering language model pre-training, sentence embedding training, and the language-understanding diagnostics, including failed and re-executed runs. We computed this total from the run durations and GPU counts recorded in Weights \& Biases, and from job logs for the runs not recorded there. Training was conducted primarily on NVIDIA A100 (40GB and 80GB) GPUs, with some runs on NVIDIA H100 and NVIDIA RTX PRO 6000 Blackwell GPUs. Hours on other GPUs are converted to A100-equivalent hours using their measured training speed.


\begin{thebibliography}{53}
\providecommand{\natexlab}[1]{#1}

\bibitem[{Ainslie et~al.(2023)Ainslie, Lee-Thorp, de~Jong, Zemlyanskiy, Lebron,
  and Sanghai}]{gqa}
Joshua Ainslie, James Lee-Thorp, Michiel de~Jong, Yury Zemlyanskiy, Federico
  Lebron, and Sumit Sanghai. 2023.
\newblock \href {https://doi.org/10.18653/v1/2023.emnlp-main.298} {{GQA}:
  Training generalized multi-query transformer models from multi-head
  checkpoints}.
\newblock In \emph{Proceedings of the 2023 Conference on Empirical Methods in
  Natural Language Processing}, pages 4895--4901, Singapore. Association for
  Computational Linguistics.

\bibitem[{Bajaj et~al.(2018)Bajaj, Campos, Craswell, Deng, Gao, Liu, Majumder,
  McNamara, Mitra, Nguyen, Rosenberg, Song, Stoica, Tiwary, and Wang}]{msmarco}
Payal Bajaj, Daniel Campos, Nick Craswell, Li~Deng, Jianfeng Gao, Xiaodong Liu,
  Rangan Majumder, Andrew McNamara, Bhaskar Mitra, Tri Nguyen, Mir Rosenberg,
  Xia Song, Alina Stoica, Saurabh Tiwary, and Tong Wang. 2018.
\newblock \href {https://arxiv.org/abs/1611.09268} {Ms marco: A human generated
  machine reading comprehension dataset}.
\newblock \emph{Preprint}, arXiv:1611.09268.

\bibitem[{BehnamGhader et~al.(2024)BehnamGhader, Adlakha, Mosbach, Bahdanau,
  Chapados, and Reddy}]{llm2vec}
Parishad BehnamGhader, Vaibhav Adlakha, Marius Mosbach, Dzmitry Bahdanau,
  Nicolas Chapados, and Siva Reddy. 2024.
\newblock \href {https://openreview.net/forum?id=IW1PR7vEBf} {{LLM2Vec}: Large
  language models are secretly powerful text encoders}.
\newblock \emph{Proceedings of the First Conference on Language Modeling}.

\bibitem[{Boizard et~al.(2026)Boizard, Deschamps-Berger, Gisserot-Boukhlef,
  Hudelot, and Colombo}]{bidirlm}
Nicolas Boizard, Th{\'e}o Deschamps-Berger, Hippolyte Gisserot-Boukhlef,
  C{\'e}line Hudelot, and Pierre Colombo. 2026.
\newblock \href {https://openreview.net/forum?id=SzcTaKW9z5} {Bidir{LM}: From
  text to omnimodal bidirectional encoders by adapting and composing causal
  {LLM}s}.
\newblock In \emph{Third Conference on Language Modeling}.

\bibitem[{Bowman et~al.(2015)Bowman, Angeli, Potts, and Manning}]{snli}
Samuel~R. Bowman, Gabor Angeli, Christopher Potts, and Christopher~D. Manning.
  2015.
\newblock \href {https://doi.org/10.18653/v1/D15-1075} {A large annotated
  corpus for learning natural language inference}.
\newblock In \emph{Proceedings of the 2015 Conference on Empirical Methods in
  Natural Language Processing}, pages 632--642, Lisbon, Portugal. Association
  for Computational Linguistics.

\bibitem[{Cer et~al.(2017)Cer, Diab, Agirre, Lopez-Gazpio, and
  Specia}]{cer-etal-2017-semeval}
Daniel Cer, Mona Diab, Eneko Agirre, I{\~n}igo Lopez-Gazpio, and Lucia Specia.
  2017.
\newblock \href {https://doi.org/10.18653/v1/S17-2001} {{S}em{E}val-2017 task
  1: Semantic textual similarity multilingual and crosslingual focused
  evaluation}.
\newblock In \emph{Proceedings of the 11th International Workshop on Semantic
  Evaluation ({S}em{E}val-2017)}, pages 1--14, Vancouver, Canada. Association
  for Computational Linguistics.

\bibitem[{Cheng et~al.(2024)Cheng, Gu, Huang, Bi, Huang, and Wei}]{instructlm}
Daixuan Cheng, Yuxian Gu, Shaohan Huang, Junyu Bi, Minlie Huang, and Furu Wei.
  2024.
\newblock \href {https://doi.org/10.18653/v1/2024.emnlp-main.148} {Instruction
  pre-training: Language models are supervised multitask learners}.
\newblock In \emph{Proceedings of the 2024 Conference on Empirical Methods in
  Natural Language Processing}, pages 2529--2550, Miami, Florida, USA.
  Association for Computational Linguistics.

\bibitem[{Dao(2024)}]{flashattention2}
Tri Dao. 2024.
\newblock \href
  {https://proceedings.iclr.cc/paper_files/paper/2024/file/98ed250b203d1ac6b24bbcf263e3d4a7-Paper-Conference.pdf}
  {{FlashAttention-2}: Faster attention with better parallelism and work
  partitioning}.
\newblock In \emph{International Conference on Learning Representations},
  volume 2024, pages 35549--35562.

\bibitem[{Devlin et~al.(2019)Devlin, Chang, Lee, and Toutanova}]{bert}
Jacob Devlin, Ming-Wei Chang, Kenton Lee, and Kristina Toutanova. 2019.
\newblock \href {https://doi.org/10.18653/v1/N19-1423} {{BERT}: Pre-training of
  deep bidirectional transformers for language understanding}.
\newblock In \emph{Proceedings of the 2019 Conference of the North {A}merican
  Chapter of the Association for Computational Linguistics: Human Language
  Technologies, Volume 1 (Long and Short Papers)}, pages 4171--4186,
  Minneapolis, Minnesota. Association for Computational Linguistics.

\bibitem[{Enevoldsen et~al.(2025)Enevoldsen, Chung, Kerboua, Kardos, Mathur,
  Stap, Gala, Siblini, Krzemi{\'n}ski, Winata, Sturua, Utpala, Ciancone,
  Schaeffer, Misra, Dhakal, Rystr{\o}m, Solomatin, {\c{C}}a{\u{g}}atan, Kundu,
  Bernstorff, Xiao, Sukhlecha, Pahwa, Po{\'s}wiata, GV, Ashraf, Auras,
  Pl{\"u}ster, Harries, Magne, Mohr, Zhu, Gisserot-Boukhlef, Aarsen, Kostkan,
  Wojtasik, Lee, Suppa, Zhang, Rocca, Hamdy, Michail, Yang, Faysse, Vatolin,
  Thakur, Dey, Vasani, Chitale, Tedeschi, Tai, Snegirev, Hendriksen,
  G{\"u}nther, Xia, Shi, Lu, Clive, K, Anna, Wehrli, Tikhonova, Panchal,
  Abramov, Ostendorff, Liu, Clematide, Miranda, Fenogenova, Song, Bin~Safi, Li,
  Borghini, Cassano, Hansen, Hooker, Xiao, Adlakha, Weller, Reddy, and
  Muennighoff}]{mmteb}
Kenneth Enevoldsen, Isaac Chung, Imene Kerboua, M{\'a}rton Kardos, Ashwin
  Mathur, David Stap, Jay Gala, Wissam Siblini, Dominik Krzemi{\'n}ski, Genta
  Winata, Saba Sturua, Saiteja Utpala, Mathieu Ciancone, Marion Schaeffer,
  Diganta Misra, Shreeya Dhakal, Jonathan Rystr{\o}m, Roman Solomatin, {\"O}mer
  {\c{C}}a{\u{g}}atan, and 63 others. 2025.
\newblock \href
  {https://proceedings.iclr.cc/paper_files/paper/2025/file/fc0e3f908a2116ba529ad0a1530a3675-Paper-Conference.pdf}
  {{MMTEB}: Massive multilingual text embedding benchmark}.
\newblock In \emph{International Conference on Learning Representations},
  volume 2025, pages 101715--101771.

\bibitem[{Gao et~al.(2021)Gao, Yao, and Chen}]{simcse}
Tianyu Gao, Xingcheng Yao, and Danqi Chen. 2021.
\newblock \href {https://doi.org/10.18653/v1/2021.emnlp-main.552} {{S}im{CSE}:
  Simple contrastive learning of sentence embeddings}.
\newblock In \emph{Proceedings of the 2021 Conference on Empirical Methods in
  Natural Language Processing}, pages 6894--6910, Online and Punta Cana,
  Dominican Republic. Association for Computational Linguistics.

\bibitem[{Garrido et~al.(2023)Garrido, Balestriero, Najman, and
  Lecun}]{garrido2023rankme}
Quentin Garrido, Randall Balestriero, Laurent Najman, and Yann Lecun. 2023.
\newblock \href {https://proceedings.mlr.press/v202/garrido23a.html} {Rankme:
  Assessing the downstream performance of pretrained self-supervised
  representations by their rank}.
\newblock In \emph{International Conference on Machine Learning}, pages
  10929--10974. PMLR.

\bibitem[{Gisserot-Boukhlef et~al.(2026)Gisserot-Boukhlef, Boizard, Faysse,
  Alves, Malherbe, Martins, Hudelot, and Colombo}]{mlmvsclm}
Hippolyte Gisserot-Boukhlef, Nicolas Boizard, Manuel Faysse, Duarte Alves,
  Emmanuel Malherbe, Andre Martins, C{\'e}line Hudelot, and Pierre Colombo.
  2026.
\newblock \href
  {https://proceedings.iclr.cc/paper_files/paper/2026/file/5b0c0b2c2efdd736a53688ebfdc3bcdb-Paper-Conference.pdf}
  {Should we still pretrain encoders with masked language modeling?}
\newblock In \emph{International Conference on Learning Representations},
  volume 2026, pages 55510--55533.

\bibitem[{Guo et~al.(2020)Guo, Dai, Vrande{\v{c}}i{\'c}, and Al-Rfou}]{wiki40b}
Mandy Guo, Zihang Dai, Denny Vrande{\v{c}}i{\'c}, and Rami Al-Rfou. 2020.
\newblock \href {https://aclanthology.org/2020.lrec-1.297/} {{W}iki-40{B}:
  Multilingual language model dataset}.
\newblock In \emph{Proceedings of the Twelfth Language Resources and Evaluation
  Conference}, pages 2440--2452, Marseille, France. European Language Resources
  Association.

\bibitem[{Hoffmann et~al.(2022)Hoffmann, Borgeaud, Mensch, Buchatskaya, Cai,
  Rutherford, de~Las~Casas, Hendricks, Welbl, Clark, Hennigan, Noland,
  Millican, van~den Driessche, Damoc, Guy, Osindero, Simonyan, Elsen, Vinyals,
  Rae, and Sifre}]{chinchilla}
Jordan Hoffmann, Sebastian Borgeaud, Arthur Mensch, Elena Buchatskaya, Trevor
  Cai, Eliza Rutherford, Diego de~Las~Casas, Lisa~Anne Hendricks, Johannes
  Welbl, Aidan Clark, Tom Hennigan, Eric Noland, Katie Millican, George van~den
  Driessche, Bogdan Damoc, Aurelia Guy, Simon Osindero, Karen Simonyan, Erich
  Elsen, and 3 others. 2022.
\newblock \href
  {https://proceedings.neurips.cc/paper_files/paper/2022/hash/c1e2faff6f588870935f114ebe04a3e5-Abstract-Conference.html}
  {Training compute-optimal large language models}.
\newblock In \emph{Proceedings of the 36th International Conference on Neural
  Information Processing Systems}, NIPS '22, Red Hook, NY, USA. Curran
  Associates Inc.

\bibitem[{Jiang et~al.(2023)Jiang, Sablayrolles, Mensch, Bamford, Chaplot,
  de~las Casas, Bressand, Lengyel, Lample, Saulnier, Lavaud, Lachaux, Stock,
  Scao, Lavril, Wang, Lacroix, and Sayed}]{mistral}
Albert~Q. Jiang, Alexandre Sablayrolles, Arthur Mensch, Chris Bamford,
  Devendra~Singh Chaplot, Diego de~las Casas, Florian Bressand, Gianna Lengyel,
  Guillaume Lample, Lucile Saulnier, Lélio~Renard Lavaud, Marie-Anne Lachaux,
  Pierre Stock, Teven~Le Scao, Thibaut Lavril, Thomas Wang, Timothée Lacroix,
  and William~El Sayed. 2023.
\newblock \href {https://arxiv.org/abs/2310.06825} {Mistral 7b}.
\newblock \emph{Preprint}, arXiv:2310.06825.

\bibitem[{Joshi et~al.(2017)Joshi, Choi, Weld, and Zettlemoyer}]{triviaqa}
Mandar Joshi, Eunsol Choi, Daniel Weld, and Luke Zettlemoyer. 2017.
\newblock \href {https://doi.org/10.18653/v1/P17-1147} {{T}rivia{QA}: A large
  scale distantly supervised challenge dataset for reading comprehension}.
\newblock In \emph{Proceedings of the 55th Annual Meeting of the Association
  for Computational Linguistics (Volume 1: Long Papers)}, pages 1601--1611,
  Vancouver, Canada. Association for Computational Linguistics.

\bibitem[{Kaplan et~al.(2020)Kaplan, McCandlish, Henighan, Brown, Chess, Child,
  Gray, Radford, Wu, and Amodei}]{kaplan2020scaling}
Jared Kaplan, Sam McCandlish, Tom Henighan, Tom~B. Brown, Benjamin Chess, Rewon
  Child, Scott Gray, Alec Radford, Jeffrey Wu, and Dario Amodei. 2020.
\newblock \href {https://arxiv.org/abs/2001.08361} {Scaling laws for neural
  language models}.
\newblock \emph{Preprint}, arXiv:2001.08361.

\bibitem[{Kurihara et~al.(2022)Kurihara, Kawahara, and Shibata}]{jglue}
Kentaro Kurihara, Daisuke Kawahara, and Tomohide Shibata. 2022.
\newblock \href {https://aclanthology.org/2022.lrec-1.317/} {{JGLUE}:
  {J}apanese general language understanding evaluation}.
\newblock In \emph{Proceedings of the Thirteenth Language Resources and
  Evaluation Conference}, pages 2957--2966, Marseille, France. European
  Language Resources Association.

\bibitem[{Kwiatkowski et~al.(2019)Kwiatkowski, Palomaki, Redfield, Collins,
  Parikh, Alberti, Epstein, Polosukhin, Devlin, Lee, Toutanova, Jones, Kelcey,
  Chang, Dai, Uszkoreit, Le, and Petrov}]{nq}
Tom Kwiatkowski, Jennimaria Palomaki, Olivia Redfield, Michael Collins, Ankur
  Parikh, Chris Alberti, Danielle Epstein, Illia Polosukhin, Jacob Devlin,
  Kenton Lee, Kristina Toutanova, Llion Jones, Matthew Kelcey, Ming-Wei Chang,
  Andrew~M. Dai, Jakob Uszkoreit, Quoc Le, and Slav Petrov. 2019.
\newblock \href {https://doi.org/10.1162/tacl_a_00276} {Natural questions: A
  benchmark for question answering research}.
\newblock \emph{Transactions of the Association for Computational Linguistics},
  7:452--466.

\bibitem[{Lee et~al.(2025)Lee, Roy, Xu, Raiman, Shoeybi, Catanzaro, and
  Ping}]{nv-embed}
Chankyu Lee, Rajarshi Roy, Mengyao Xu, Jonathan Raiman, Mohammad Shoeybi, Bryan
  Catanzaro, and Wei Ping. 2025.
\newblock \href
  {https://proceedings.iclr.cc/paper_files/paper/2025/file/c4bf73386022473a652a18941e9ea6f8-Paper-Conference.pdf}
  {{NV-Embed}: Improved techniques for training {LLM}s as generalist embedding
  models}.
\newblock In \emph{International Conference on Learning Representations},
  volume 2025, pages 79310--79333.

\bibitem[{Lewis et~al.(2020)Lewis, Perez, Piktus, Petroni, Karpukhin, Goyal,
  K\"{u}ttler, Lewis, Yih, Rockt\"{a}schel, Riedel, and Kiela}]{rag}
Patrick Lewis, Ethan Perez, Aleksandra Piktus, Fabio Petroni, Vladimir
  Karpukhin, Naman Goyal, Heinrich K\"{u}ttler, Mike Lewis, Wen-tau Yih, Tim
  Rockt\"{a}schel, Sebastian Riedel, and Douwe Kiela. 2020.
\newblock \href
  {https://proceedings.neurips.cc/paper/2020/hash/6b493230205f780e1bc26945df7481e5-Abstract.html}
  {Retrieval-augmented generation for knowledge-intensive nlp tasks}.
\newblock In \emph{Proceedings of the 34th International Conference on Neural
  Information Processing Systems}, NIPS '20, Red Hook, NY, USA. Curran
  Associates Inc.

\bibitem[{Li et~al.(2026)Li, Ohagi, Ri, Fukuchi, Shibata, and Kawahara}]{jmteb}
Shengzhe Li, Masaya Ohagi, Ryokan Ri, Akihiko Fukuchi, Tomohide Shibata, and
  Daisuke Kawahara. 2026.
\newblock \href {https://doi.org/10.63317/5ouzpv2f2f6k} {Jmteb and jmteb-lite:
  Japanese massive text embedding benchmark and its lightweight version}.
\newblock In \emph{Proceedings of the Fifteenth Language Resources and
  Evaluation Conference (LREC 2026)}, pages 7423--7434, Palma, Mallorca, Spain.
  European Language Resources Association (ELRA).

\bibitem[{Li et~al.(2023)Li, Zhang, Zhang, Long, Xie, and Zhang}]{gte}
Zehan Li, Xin Zhang, Yanzhao Zhang, Dingkun Long, Pengjun Xie, and Meishan
  Zhang. 2023.
\newblock \href {https://arxiv.org/abs/2308.03281} {Towards general text
  embeddings with multi-stage contrastive learning}.
\newblock \emph{Preprint}, arXiv:2308.03281.

\bibitem[{Liu et~al.(2024)Liu, Li, Xiao, Shao, and Lian}]{llama2vec}
Zheng Liu, Chaofan Li, Shitao Xiao, Yingxia Shao, and Defu Lian. 2024.
\newblock \href {https://doi.org/10.18653/v1/2024.acl-long.191} {{L}lama2{V}ec:
  Unsupervised adaptation of large language models for dense retrieval}.
\newblock In \emph{Proceedings of the 62nd Annual Meeting of the Association
  for Computational Linguistics (Volume 1: Long Papers)}, pages 3490--3500,
  Bangkok, Thailand. Association for Computational Linguistics.

\bibitem[{Muennighoff(2022)}]{sgpt}
Niklas Muennighoff. 2022.
\newblock \href {https://arxiv.org/abs/2202.08904} {{SGPT}: {GPT} {S}entence
  {E}mbeddings for {S}emantic {S}earch}.
\newblock \emph{Preprint}, arXiv:2202.08904.

\bibitem[{Muennighoff et~al.(2023)Muennighoff, Tazi, Magne, and Reimers}]{mteb}
Niklas Muennighoff, Nouamane Tazi, Loic Magne, and Nils Reimers. 2023.
\newblock \href {https://doi.org/10.18653/v1/2023.eacl-main.148} {{MTEB}:
  Massive text embedding benchmark}.
\newblock In \emph{Proceedings of the 17th Conference of the European Chapter
  of the Association for Computational Linguistics}, pages 2014--2037,
  Dubrovnik, Croatia. Association for Computational Linguistics.

\bibitem[{Penedo et~al.(2025)Penedo, Kydl{\'\i}{\v{c}}ek, Sabol{\v{c}}ec,
  Messmer, Foroutan, Kargaran, Raffel, Jaggi, Werra, and
  Wolf}]{penedo2025fineweb}
Guilherme Penedo, Hynek Kydl{\'\i}{\v{c}}ek, Vinko Sabol{\v{c}}ec, Bettina
  Messmer, Negar Foroutan, Amir~Hossein Kargaran, Colin Raffel, Martin Jaggi,
  Leandro~Von Werra, and Thomas Wolf. 2025.
\newblock \href {https://openreview.net/forum?id=jnRBe6zatP} {Fineweb2: One
  pipeline to scale them all {\textemdash} adapting pre-training data
  processing to every language}.
\newblock In \emph{Second Conference on Language Modeling}.

\bibitem[{Penedo et~al.(2024)Penedo, Kydl\'{\i}\v{c}ek, allal, Lozhkov,
  Mitchell, Raffel, Von~Werra, and Wolf}]{fineweb}
Guilherme Penedo, Hynek Kydl\'{\i}\v{c}ek, Loubna~Ben allal, Anton Lozhkov,
  Margaret Mitchell, Colin Raffel, Leandro Von~Werra, and Thomas Wolf. 2024.
\newblock \href {https://doi.org/10.52202/079017-0970} {The {FineWeb} datasets:
  Decanting the web for the finest text data at scale}.
\newblock In \emph{Advances in Neural Information Processing Systems},
  volume~37, pages 30811--30849. Curran Associates, Inc.

\bibitem[{Rajpurkar et~al.(2016)Rajpurkar, Zhang, Lopyrev, and Liang}]{squad}
Pranav Rajpurkar, Jian Zhang, Konstantin Lopyrev, and Percy Liang. 2016.
\newblock \href {https://doi.org/10.18653/v1/D16-1264} {{SQ}u{AD}: 100,000+
  questions for machine comprehension of text}.
\newblock In \emph{Proceedings of the 2016 Conference on Empirical Methods in
  Natural Language Processing}, pages 2383--2392, Austin, Texas. Association
  for Computational Linguistics.

\bibitem[{Reimers and Gurevych(2019)}]{sbert}
Nils Reimers and Iryna Gurevych. 2019.
\newblock \href {https://doi.org/10.18653/v1/D19-1410} {Sentence-{BERT}:
  Sentence embeddings using {S}iamese {BERT}-networks}.
\newblock In \emph{Proceedings of the 2019 Conference on Empirical Methods in
  Natural Language Processing and the 9th International Joint Conference on
  Natural Language Processing (EMNLP-IJCNLP)}, pages 3982--3992, Hong Kong,
  China. Association for Computational Linguistics.

\bibitem[{Rudman et~al.(2023)Rudman, Chen, and Eickhoff}]{rudman2023outlier}
William Rudman, Catherine Chen, and Carsten Eickhoff. 2023.
\newblock \href {https://doi.org/10.18653/v1/2023.emnlp-main.901} {Outlier
  dimensions encode task specific knowledge}.
\newblock In \emph{Proceedings of the 2023 Conference on Empirical Methods in
  Natural Language Processing}, pages 14596--14605, Singapore. Association for
  Computational Linguistics.

\bibitem[{Sakai et~al.(2024)Sakai, Nohejl, Hang, Kamigaito, and
  Watanabe}]{sakai2024templates}
Yusuke Sakai, Adam Nohejl, Jiangnan Hang, Hidetaka Kamigaito, and Taro
  Watanabe. 2024.
\newblock \href {https://doi.org/10.18653/v1/2024.blackboxnlp-1.31} {Toward the
  evaluation of large language models considering score variance across
  instruction templates}.
\newblock In \emph{Proceedings of the 7th BlackboxNLP Workshop: Analyzing and
  Interpreting Neural Networks for NLP}, pages 499--529, Miami, Florida, US.
  Association for Computational Linguistics.

\bibitem[{Someya et~al.(2024)Someya, Sugimoto, and Oseki}]{jcola}
Taiga Someya, Yushi Sugimoto, and Yohei Oseki. 2024.
\newblock \href {https://aclanthology.org/2024.lrec-main.828/} {{JC}o{LA}:
  {J}apanese corpus of linguistic acceptability}.
\newblock In \emph{Proceedings of the 2024 Joint International Conference on
  Computational Linguistics, Language Resources and Evaluation (LREC-COLING
  2024)}, pages 9477--9488, Torino, Italia. ELRA and ICCL.

\bibitem[{Su et~al.(2023)Su, Shi, Kasai, Wang, Hu, Ostendorf, Yih, Smith,
  Zettlemoyer, and Yu}]{instructor}
Hongjin Su, Weijia Shi, Jungo Kasai, Yizhong Wang, Yushi Hu, Mari Ostendorf,
  Wen-tau Yih, Noah~A. Smith, Luke Zettlemoyer, and Tao Yu. 2023.
\newblock \href {https://doi.org/10.18653/v1/2023.findings-acl.71} {One
  embedder, any task: Instruction-finetuned text embeddings}.
\newblock In \emph{Findings of the Association for Computational Linguistics:
  ACL 2023}, pages 1102--1121, Toronto, Canada. Association for Computational
  Linguistics.

\bibitem[{Talmor et~al.(2019)Talmor, Herzig, Lourie, and
  Berant}]{commonsenseqa}
Alon Talmor, Jonathan Herzig, Nicholas Lourie, and Jonathan Berant. 2019.
\newblock \href {https://doi.org/10.18653/v1/N19-1421} {{C}ommonsense{QA}: A
  question answering challenge targeting commonsense knowledge}.
\newblock In \emph{Proceedings of the 2019 Conference of the North {A}merican
  Chapter of the Association for Computational Linguistics: Human Language
  Technologies, Volume 1 (Long and Short Papers)}, pages 4149--4158,
  Minneapolis, Minnesota. Association for Computational Linguistics.

\bibitem[{Timkey and van Schijndel(2021)}]{timkey2021rogue}
William Timkey and Marten van Schijndel. 2021.
\newblock \href {https://doi.org/10.18653/v1/2021.emnlp-main.372} {All bark and
  no bite: Rogue dimensions in transformer language models obscure
  representational quality}.
\newblock In \emph{Proceedings of the 2021 Conference on Empirical Methods in
  Natural Language Processing}, pages 4527--4546, Online and Punta Cana,
  Dominican Republic. Association for Computational Linguistics.

\bibitem[{Touvron et~al.(2023)Touvron, Lavril, Izacard, Martinet, Lachaux,
  Lacroix, Rozière, Goyal, Hambro, Azhar, Rodriguez, Joulin, Grave, and
  Lample}]{llama}
Hugo Touvron, Thibaut Lavril, Gautier Izacard, Xavier Martinet, Marie-Anne
  Lachaux, Timothée Lacroix, Baptiste Rozière, Naman Goyal, Eric Hambro,
  Faisal Azhar, Aurelien Rodriguez, Armand Joulin, Edouard Grave, and Guillaume
  Lample. 2023.
\newblock \href {https://arxiv.org/abs/2302.13971} {Llama: Open and efficient
  foundation language models}.
\newblock \emph{Preprint}, arXiv:2302.13971.

\bibitem[{Tsukagoshi and Sasano(2026)}]{ruri}
Hayato Tsukagoshi and Ryohei Sasano. 2026.
\newblock \href {https://doi.org/10.5715/jnlp.33.591} {Ruri: Japanese general
  text embeddings}.
\newblock \emph{Journal of Natural Language Processing}, 33(2):591--629.

\bibitem[{van~den Oord et~al.(2019)van~den Oord, Li, and Vinyals}]{infonce}
Aaron van~den Oord, Yazhe Li, and Oriol Vinyals. 2019.
\newblock \href {https://arxiv.org/abs/1807.03748} {Representation learning
  with contrastive predictive coding}.
\newblock \emph{Preprint}, arXiv:1807.03748.

\bibitem[{Wang et~al.(2024{\natexlab{a}})Wang, Yang, Huang, Jiao, Yang, Jiang,
  Majumder, and Wei}]{e5}
Liang Wang, Nan Yang, Xiaolong Huang, Binxing Jiao, Linjun Yang, Daxin Jiang,
  Rangan Majumder, and Furu Wei. 2024{\natexlab{a}}.
\newblock \href {https://arxiv.org/abs/2212.03533} {Text embeddings by
  weakly-supervised contrastive pre-training}.
\newblock \emph{Preprint}, arXiv:2212.03533.

\bibitem[{Wang et~al.(2024{\natexlab{b}})Wang, Yang, Huang, Yang, Majumder, and
  Wei}]{e5mistral}
Liang Wang, Nan Yang, Xiaolong Huang, Linjun Yang, Rangan Majumder, and Furu
  Wei. 2024{\natexlab{b}}.
\newblock \href {https://doi.org/10.18653/v1/2024.acl-long.642} {Improving text
  embeddings with large language models}.
\newblock In \emph{Proceedings of the 62nd Annual Meeting of the Association
  for Computational Linguistics (Volume 1: Long Papers)}, pages 11897--11916,
  Bangkok, Thailand. Association for Computational Linguistics.

\bibitem[{Wang and Isola(2020)}]{understanding}
Tongzhou Wang and Phillip Isola. 2020.
\newblock \href {https://proceedings.mlr.press/v119/wang20k.html}
  {Understanding contrastive representation learning through alignment and
  uniformity on the hypersphere}.
\newblock In \emph{Proceedings of the 37th International Conference on Machine
  Learning}, ICML'20. JMLR.org.

\bibitem[{Warner et~al.(2025)Warner, Chaffin, Clavi{\'e}, Weller,
  Hallstr{\"o}m, Taghadouini, Gallagher, Biswas, Ladhak, Aarsen, Adams, Howard,
  and Poli}]{modernbert}
Benjamin Warner, Antoine Chaffin, Benjamin Clavi{\'e}, Orion Weller, Oskar
  Hallstr{\"o}m, Said Taghadouini, Alexis Gallagher, Raja Biswas, Faisal
  Ladhak, Tom Aarsen, Griffin~Thomas Adams, Jeremy Howard, and Iacopo Poli.
  2025.
\newblock \href {https://doi.org/10.18653/v1/2025.acl-long.127} {Smarter,
  better, faster, longer: A modern bidirectional encoder for fast, memory
  efficient, and long context finetuning and inference}.
\newblock In \emph{Proceedings of the 63rd Annual Meeting of the Association
  for Computational Linguistics (Volume 1: Long Papers)}, pages 2526--2547,
  Vienna, Austria. Association for Computational Linguistics.

\bibitem[{Warstadt et~al.(2019)Warstadt, Singh, and Bowman}]{cola}
Alex Warstadt, Amanpreet Singh, and Samuel~R. Bowman. 2019.
\newblock \href {https://doi.org/10.1162/tacl_a_00290} {Neural network
  acceptability judgments}.
\newblock \emph{Transactions of the Association for Computational Linguistics},
  7:625--641.

\bibitem[{Weller et~al.(2026)Weller, Ricci, Marone, Chaffin, Lawrie, and
  Van~Durme}]{ettin}
Orion Weller, Kathryn Ricci, Marc Marone, Antoine Chaffin, Dawn Lawrie, and Ben
  Van~Durme. 2026.
\newblock \href
  {https://proceedings.iclr.cc/paper_files/paper/2026/file/e78aee320c3467fcb3c5948547d9851d-Paper-Conference.pdf}
  {{Seq} vs {Seq}: An open suite of paired encoders and decoders}.
\newblock In \emph{International Conference on Learning Representations},
  volume 2026, pages 143110--143133.

\bibitem[{Williams et~al.(2018)Williams, Nangia, and
  Bowman}]{williams-etal-2018-broad}
Adina Williams, Nikita Nangia, and Samuel~R. Bowman. 2018.
\newblock \href {https://doi.org/10.18653/v1/N18-1101} {A broad-coverage
  challenge corpus for sentence understanding through inference}.
\newblock In \emph{Proceedings of the 2018 Conference of the North {A}merican
  Chapter of the Association for Computational Linguistics: Human Language
  Technologies, Volume 1 (Long Papers)}, pages 1112--1122, New Orleans,
  Louisiana. Association for Computational Linguistics.

\bibitem[{Wolf et~al.(2020)Wolf, Debut, Sanh, Chaumond, Delangue, Moi, Cistac,
  Rault, Louf, Funtowicz, Davison, Shleifer, von Platen, Ma, Jernite, Plu, Xu,
  Le~Scao, Gugger, Drame, Lhoest, and Rush}]{transformerslib}
Thomas Wolf, Lysandre Debut, Victor Sanh, Julien Chaumond, Clement Delangue,
  Anthony Moi, Pierric Cistac, Tim Rault, Remi Louf, Morgan Funtowicz, Joe
  Davison, Sam Shleifer, Patrick von Platen, Clara Ma, Yacine Jernite, Julien
  Plu, Canwen Xu, Teven Le~Scao, Sylvain Gugger, and 3 others. 2020.
\newblock \href {https://doi.org/10.18653/v1/2020.emnlp-demos.6} {Transformers:
  State-of-the-art natural language processing}.
\newblock In \emph{Proceedings of the 2020 Conference on Empirical Methods in
  Natural Language Processing: System Demonstrations}, pages 38--45, Online.
  Association for Computational Linguistics.

\bibitem[{Wunderle et~al.(2026)Wunderle, Ehrmanntraut, Pfister, Jannidis, and
  Hotho}]{moderngbert}
Julia Wunderle, Anton Ehrmanntraut, Jan Pfister, Fotis Jannidis, and Andreas
  Hotho. 2026.
\newblock \href {https://doi.org/10.63317/4s26zh4i323y} {New encoders for
  {G}erman trained from scratch: Comparing {ModernGBERT} with converted
  {LLM2Vec} models}.
\newblock In \emph{Proceedings of the Fifteenth Language Resources and
  Evaluation Conference ({LREC} 2026)}, pages 10424--10446, Palma, Mallorca,
  Spain. European Language Resources Association (ELRA).

\bibitem[{Xiao et~al.(2024{\natexlab{a}})Xiao, Tian, Chen, Han, and
  Lewis}]{xiao2024efficient}
Guangxuan Xiao, Yuandong Tian, Beidi Chen, Song Han, and Mike Lewis.
  2024{\natexlab{a}}.
\newblock \href
  {https://proceedings.iclr.cc/paper_files/paper/2024/file/5e5fd18f863cbe6d8ae392a93fd271c9-Paper-Conference.pdf}
  {Efficient streaming language models with attention sinks}.
\newblock In \emph{International Conference on Learning Representations},
  volume 2024, pages 21875--21895.

\bibitem[{Xiao et~al.(2024{\natexlab{b}})Xiao, Liu, Zhang, Muennighoff, Lian,
  and Nie}]{cpack}
Shitao Xiao, Zheng Liu, Peitian Zhang, Niklas Muennighoff, Defu Lian, and
  Jian-Yun Nie. 2024{\natexlab{b}}.
\newblock \href {https://doi.org/10.1145/3626772.3657878} {C-pack: Packed
  resources for general chinese embeddings}.
\newblock In \emph{Proceedings of the 47th International ACM SIGIR Conference
  on Research and Development in Information Retrieval}, SIGIR '24, page
  641–649, New York, NY, USA. Association for Computing Machinery.

\bibitem[{Yang et~al.(2025)Yang, Yang, Zhang, Hui, Zheng, Yu, Li, Liu, Huang,
  Wei, Lin, Yang, Tu, Zhang, Yang, Yang, Zhou, Lin, Dang, Lu, Bao, Yang, Yu,
  Li, Xue, Zhang, Zhu, Men, Lin, Li, Tang, Xia, Ren, Ren, Fan, Su, Zhang, Wan,
  Liu, Cui, Zhang, and Qiu}]{qwen25}
An~Yang, Baosong Yang, Beichen Zhang, Binyuan Hui, Bo~Zheng, Bowen Yu,
  Chengyuan Li, Dayiheng Liu, Fei Huang, Haoran Wei, Huan Lin, Jian Yang,
  Jianhong Tu, Jianwei Zhang, Jianxin Yang, Jiaxi Yang, Jingren Zhou, Junyang
  Lin, Kai Dang, and 23 others. 2025.
\newblock \href {https://arxiv.org/abs/2412.15115} {Qwen2.5 technical report}.
\newblock \emph{Preprint}, arXiv:2412.15115.

\bibitem[{Zhang et~al.(2023)Zhang, Thakur, Ogundepo, Kamalloo, Alfonso-Hermelo,
  Li, Liu, Rezagholizadeh, and Lin}]{miracl}
Xinyu Zhang, Nandan Thakur, Odunayo Ogundepo, Ehsan Kamalloo, David
  Alfonso-Hermelo, Xiaoguang Li, Qun Liu, Mehdi Rezagholizadeh, and Jimmy Lin.
  2023.
\newblock \href {https://doi.org/10.1162/tacl_a_00595} {{MIRACL}: A
  multilingual retrieval dataset covering 18 diverse languages}.
\newblock \emph{Transactions of the Association for Computational Linguistics},
  11:1114--1131.

\end{thebibliography}
\end{document}